\documentclass[letterpaper]{article} 
\usepackage[preprint]{aaai2027}  
\usepackage[hyphens]{url}  
\usepackage{graphicx} 
\usepackage{natbib}  
\usepackage{caption} 
\usepackage{algorithm}
\usepackage{booktabs}
\usepackage{amsmath}
\usepackage{amssymb}
\usepackage{algpseudocode}
\usepackage{newfloat}
\usepackage{listings}
\DeclareCaptionStyle{ruled}{labelfont=normalfont,labelsep=colon,strut=off} 
\floatstyle{ruled}
\newfloat{listing}{tb}{lst}{}
\floatname{listing}{Listing}

\usepackage{booktabs}

\title{SparseKAN: Compressing Kolmogorov--Arnold Networks Across Basis Functions, Neurons, and Bits}

\author{
    Kazi Ahmed Asif Fuad\textsuperscript{\rm 1},
    Lizhong Chen\textsuperscript{\rm 1}
}

\affiliations{
    \textsuperscript{\rm 1}Department of EECS\\
    Oregon State University\\
    Corvallis, OR 97331\\
    fuadk@oregonstate.edu, chenliz@oregonstate.edu
}

\begin{document}

\maketitle

\begin{abstract}
Kolmogorov--Arnold Networks (KANs) replace scalar edge weights with learnable univariate functions parameterized by multiple basis coefficients. This introduces a source of redundancy that conventional neural-network compression does not directly expose. We present \textbf{SparseKAN}, a unified approach that compresses KANs along three complementary axes: \emph{basis functions}, \emph{neurons/channels}, and \emph{numerical precision}. SparseKAN equips the base branch, nonlinear basis branch, and individual basis terms with hierarchical learnable gates trained under a differentiable active-cost objective. The learned importance structure is subsequently hardened under explicit basis and width budgets, recovered in full or low precision, and physically compacted into smaller dense tensors rather than retained as sparse masks. Experiments on MNIST, CIFAR-10, and CIFAR-100 across spline, polynomial, RBF, wavelet, and convolutional KAN variants show that the structural axes compose predictably in cost. We also find strong basis-dependent differences in term importance: coefficient-based selection outperforms matched low-order truncation by up to 15.25 accuracy points in the evaluated Gram-polynomial settings. Eight-bit quantization is broadly robust, whereas 4-bit convolutional KANs require quantization-aware adaptation. Physical compaction removes up to 73.0\% of parameters without accuracy loss on MNIST and reduces large-batch CUDA latency to as little as $0.51\times$ dense execution. On a ZCU104 FPGA, the resulting sparse low-bit models achieve up to $23.63\times$ lower inference latency, demonstrating that SparseKAN converts functional redundancy into measurable software and hardware efficiency. The SparseKAN implementation is available at
\url{https://github.com/OSU-STARLAB/SparseKAN}.
\end{abstract}


\section{Introduction}
\label{sec:introduction}

Kolmogorov--Arnold Networks (KANs) replace the scalar weight on an MLP connection with a learnable univariate function~\citep{liu2025kan}. In spline-based KANs, each edge combines a residual base activation with a basis expansion, so one connection carries multiple coefficients rather than a scalar. This added functional capacity is also a source of cost. For input width $d_{\mathrm{in}}$, output width $d_{\mathrm{out}}$, and $K$ basis functions per edge, the dominant coefficient tensor scales as $d_{\mathrm{in}}d_{\mathrm{out}}K$. KAN redundancy can therefore occur not only \emph{across} edges and neurons, but also \emph{inside} each learned function.

Standard pruning granularities miss this. A retained edge may need only a subset of its basis functions, while deleting the whole edge can remove useful computation together with redundant terms. The surviving structure may also tolerate lower precision. KAN compression thus exposes three complementary axes: \emph{basis functions}, \emph{neurons/channels}, and \emph{precision/bits}. Existing work addresses important parts of this space through efficient basis parameterizations~\citep{li2024fastkan,sidharth2024chebykan, bozorgasl2024wavkan,merin2026ltbskan,poole2025pkan}, coefficient sharing or vector quantization~\citep{raffel2025metacluster}, low-precision inference~\citep{fuad2025quantkan,errabii2026kantize}, learned architectural sparsity~\citep{bagrow2026optimized}, and specialized hardware~\citep{hoang2026kanele,errabii2026kansas,ou2025pdrkan, ou2026vikin,sudarshan2026cim}. However, what remains less explored is a general mechanism that exposes the internal basis dimension, coordinates it with width and precision, and converts the resulting structure into physically smaller executable tensors.

We introduce \textbf{SparseKAN}, a basis-aware compression framework for this purpose. SparseKAN attaches hierarchical gates to the base branch, nonlinear basis branch, and individual basis terms of each KAN edge, and optimizes them with a normalized active-cost objective. The soft-gating stage is used mainly for \emph{structure discovery and adaptation}; explicit compression is imposed during hardening through a basis and a neuron/channel budget. The surviving model is then recovered in full precision or with low-bit quantization-aware training (QAT). When supports are structurally compatible, SparseKAN gathers retained basis terms and slices dead hidden dimensions, turning learned sparsity into smaller dense tensors rather than zeros in the original model.

The experiments show why these axes should remain distinct. Basis and width reductions compose predictably: on MNIST EfficientKAN, factors $0.556$ and $0.4651$ predict active cost $0.2585$, versus $0.2584$ measured. Basis identity is also family-dependent. In the evaluated Gram-polynomial settings, coefficient-based selection exceeds matched low-order truncation by up to $15.25$ accuracy points, whereas the tested B-spline models are nearly insensitive to support identity. Precision introduces a different boundary: 8-bit QAT is broadly benign, while naive 4-bit PTQ falls to $50.19\pm11.78\%$ on CIFAR-10 and $9.04\pm2.43\%$ on CIFAR-100; QAT restores useful low-bit operating points. Our five-seed audit further shows joint gate/QAT training is a robust single-pipeline option, not a universal accuracy improvement over sufficiently recovered staged compression.

Crucially, SparseKAN evaluates whether sparsity survives contact with the system. Across 54 compacted checkpoints, top-1 agreement with the corresponding masked model is at least $0.9997$. Physical compaction removes
up to $73.0\%$ of parameters without degradation on MNIST, and compact convolutional models reduce large-batch CUDA latency to as little as $0.51\times$ dense, whereas masked models remain near dense latency. On a ZCU104 HLS implementation, sparse 4-bit EfficientKAN and Gram-polynomial MLPs reach $23.63\times$ and $18.21\times$ lower inference latency, respectively; sparse 4-bit KAGN-conv reduces DSP occupancy from $96.2\%$ to $29.1\%$ for  $0.09$ accuracy points. Our contributions are as follows.

\begin{itemize}
    \item We formulate KAN compression along three native axes: basis functions, neurons/channels, and precision. We provide a common cost-gated interface for exposing their importance across multiple KAN parameterizations.
    \item We introduce per-edge and shared basis hardening and combine it with width reduction so compatible supports can be physically compacted from $d_{\ell-1}d_{\ell}K_{\ell}$ to $d'_{\ell-1}d'_{\ell}K'_{\ell}$, rather than merely masked.
    \item We validate the resulting operating points across MNIST, CIFAR-10, and CIFAR-100 using matched pruning baselines, five-seed comparative audits, physical checkpoint size, CUDA latency, and an 18-design FPGA/HLS study. The result is a basis-aware path from learned redundancy to compact, low-bit, executable KANs.
\end{itemize}

\section{Related Work}
\label{sec:related_work}

\paragraph{KAN architectures and efficient parameterizations.}
KANs originally parameterize edge functions with B-splines~\citep{liu2025kan}. Later variants replace or restructure this basis: FastKAN uses Gaussian radial basis functions~\citep{li2024fastkan}, ChebyKAN uses Chebyshev polynomials~\citep{sidharth2024chebykan}, Wav-KAN uses wavelets~\citep{bozorgasl2024wavkan}, and KAN convolutions extend such representations to vision models~\citep{drokin2024kanconv}. LSS-SKAN, LTBs-KAN, and P-KAN instead reduce cost through simpler or lower-dimensional functional parameterizations~\citep{chen2024lssskan,merin2026ltbskan,poole2025pkan}, while recent differentiable architecture optimization learns compact KAN structures directly~\citep{bagrow2026optimized}. SparseKAN is complementary: given a KAN family, it asks how much of its available basis, width, and precision is actually required.

\paragraph{KAN compression.}
The original KAN already uses regularization and pruning for model simplification~\citep{liu2025kan}. MetaCluster clusters coefficient vectors~\citep{raffel2025metacluster}, SHARe-KAN applies post-training vector quantization~\citep{smith2026sharekan}, and QuantKAN and KANtize study low-precision KAN inference~\citep{fuad2025quantkan,errabii2026kantize}. SparseKAN differs in structural granularity: it can remove individual basis terms, regularize them into shared supports, and combine that basis dimension with neuron/channel pruning and QAT. Our matched-cost edge-$L_0$ results are therefore intentionally a scope condition: basis sparsity is not universally more accurate than edge sparsity; its value is finer functional control and compactable structure.

\paragraph{Joint neural compression.}
Pruning and quantization are well established for conventional networks. Deep Compression combines pruning and quantization~\citep{han2016deepcompression}; DJPQ jointly optimizes structured pruning and mixed precision~\citep{wang2020djpq}; and Bayesian Bits unifies pruning with mixed-precision selection~\citep{vanbaalen2020bayesianbits}. SparseKAN also draws on differentiable $L_0$ gating and structured channel pruning~\citep{louizos2018l0,liu2017networkslimming}, but applies these ideas to a dimension absent from an ordinary MLP: the basis terms inside a functional edge.

\paragraph{KAN hardware and systems.}
KAN accelerators include FPGA LUT mapping with pruning and quantization~\citep{hoang2026kanele}, systolic arrays~\citep{errabii2026kansas}, reconfigurable sparse pipelines~\citep{ou2025pdrkan,ou2026vikin}, and compute-in-memory designs~\citep{sudarshan2026cim}. FlashKAT further shows that memory traffic and implementation details can dominate nominal KAN arithmetic cost~\citep{raffel2026flashkat}, and surveys of sparse transformer accelerators reach a similar conclusion: unstructured zeros rarely translate into hardware gains without explicit architectural support~\citep{fuad2023transformersparsity}. These results motivate our separation of active cost, physical model size, and measured execution cost: SparseKAN compacts compatible supports so hardware receives a smaller dense problem rather than a dense problem containing zeros.

\section{SparseKAN}
\label{sec:method}

\begin{figure*}[!ht]
    \centering
    \includegraphics[width=0.98\textwidth]{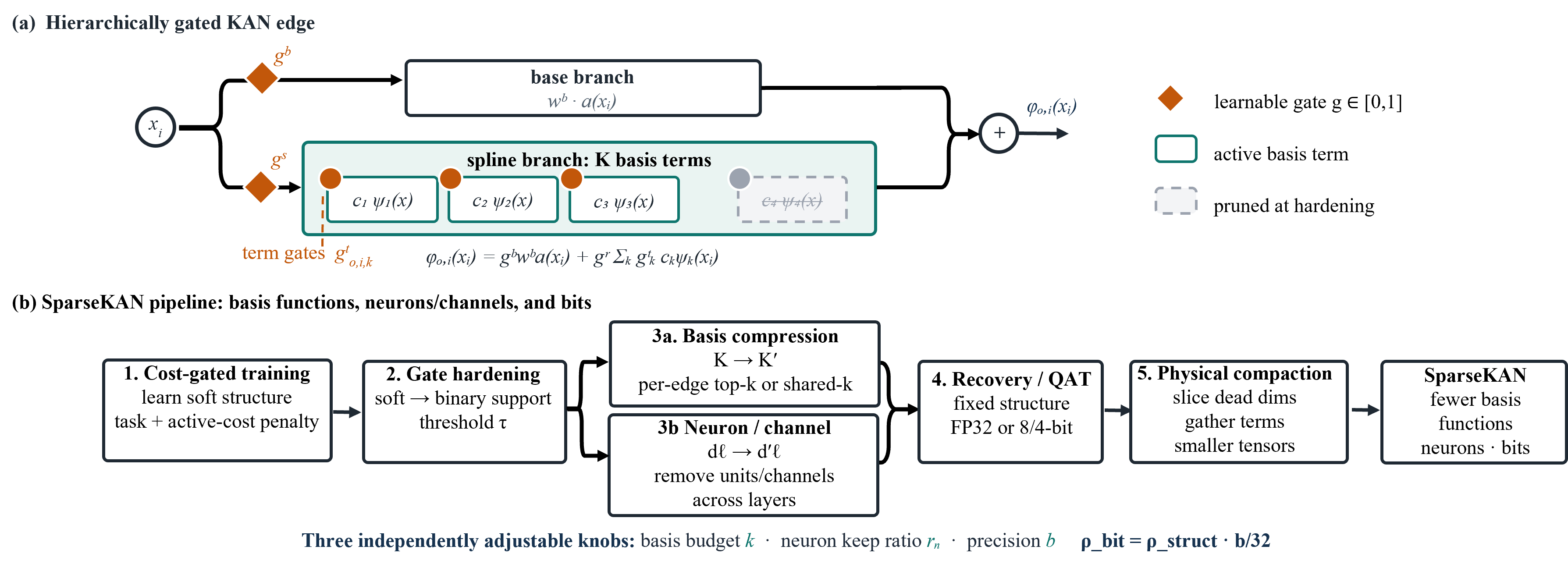}
    \caption{
    \textbf{SparseKAN compresses KANs along three complementary axes.}
    (a) Each edge contains a base branch and a basis branch,
    controlled by base, branch, and term-level gates.
    (b) Cost-gated training first exposes redundancy; hardening then reduces
    basis functions and hidden units/channels. The surviving
    model is adapted in full precision or with low-bit QAT, and structured
    supports are compacted into smaller tensors. The result uses
    fewer \emph{basis functions}, fewer \emph{neurons}, and fewer \emph{bits}.
    }
    \label{fig:method}
\end{figure*}

\subsection{From Functional Redundancy to Structured Compression}
\label{sec:method_overview}

Unlike an MLP connection, which carries a scalar weight, a KAN edge represents a learnable univariate function using multiple basis coefficients. This expressiveness creates redundancy within basis expansions, across hidden units/channels, and in the precision of the surviving parameters. SparseKAN targets these three sources jointly, as illustrated in Figure~\ref{fig:method}.

The central idea is to separate \emph{structure discovery} from \emph{structure realization}. We first train differentiable gates that expose the relative importance of the computation inside each KAN edge. The learned soft structure is then converted into explicit compression along two structural dimensions (basis functions and neurons/channels), and the remaining parameters are quantized. Structured masks can finally be physically compacted so that the selected sparsity changes the model dimensions rather than merely introducing zeros.

\subsection{Cost-Gated KANs}
\label{sec:cost_gated}

Consider edge $(i\!\rightarrow\!o)$ of a KAN layer with $K$ basis functions.
We write its gated form as:

\begin{equation}
\widetilde{\phi}_{o,i}(x_i)
=
g^{\mathrm b}_{o,i}w^{\mathrm b}_{o,i}a(x_i)
+
g^{\mathrm r}_{o,i}
\sum_{k=1}^{K}
g^{\mathrm t}_{o,i,k}
c_{o,i,k}\psi_k(x_i),
\label{eq:gated_edge}
\end{equation}
where $a(\cdot)$ is the base activation, $\psi_k(\cdot)$ is the $k$-th basis function, and $c_{o,i,k}$ is its coefficient. The gates $g^{\mathrm b}$, $g^{\mathrm r}$, and $g^{\mathrm t}$ control the
base branch, nonlinear basis branch, and individual basis terms, respectively. This formulation is basis-agnostic and applies to the spline, polynomial, RBF, wavelet, and convolutional KAN families considered in our experiments. We use sigmoid-relaxed learnable gates by default; alternative relaxations are described in Appendix~\ref{app:gates}.

To bias the model toward cheaper structure, we optimize a normalized active-cost penalty,
\begin{equation}
\mathcal{R}_{\mathrm{cost}}
=
\frac{
\lambda_{\mathrm b}C_{\mathrm b}\sum g^{\mathrm b}
+
\lambda_{\mathrm r}C_{\mathrm r}\sum g^{\mathrm r}
+
\lambda_{\mathrm t}C_{\mathrm t}
\sum g^{\mathrm r}g^{\mathrm t}
}{
C_{\mathrm b}N_{\mathrm b}
+
C_{\mathrm r}N_{\mathrm r}
+
C_{\mathrm t}N_{\mathrm t}
},
\label{eq:active_cost}
\end{equation}
where $C_{\bullet}$ and $N_{\bullet}$ denote the relative cost and dense count of each component type. The product $g^{\mathrm r}g^{\mathrm t}$ reflects the hierarchy of Eq.~\eqref{eq:gated_edge}: a term contributes only when its parent basis branch is active. Training minimizes
\begin{equation}
\mathcal{L}
=
\mathcal{L}_{\mathrm{task}}
+
\mathcal{R}_{\mathrm{cost}}
+
\lambda_{\mathrm H}\mathcal{R}_{\mathrm H},
\label{eq:training_obj}
\end{equation}
with an optional entropy penalty $\mathcal{R}_{\mathrm H}$ to encourage decisive gates. Sparsity pressure is introduced after a warm-up and increased gradually. Under our default setting this stage primarily learns and adapts the model to a useful importance structure; the desired compression level is imposed explicitly during hardening. The complete schedule and a direct cost-budget formulation are given in Appendix~\ref{app:cost}.

\subsection{Two Structural Axes: Basis Functions and Neurons}
\label{sec:structural_compression}

\paragraph{Basis-function sparsification.}
SparseKAN first reduces redundancy \emph{within} functional edges. After cost-gated training, our default term score is
\begin{equation}
s_{o,i,k}
=
|c_{o,i,k}|\,g^{\mathrm t}_{o,i,k}.
\label{eq:term_score}
\end{equation}
A flexible per-edge top-$k$ rule retains the $k$ highest-scoring terms of each active edge. Different edges may therefore use different basis subsets. Because this creates a ragged support, we additionally introduce \emph{shared-$k$} structure: all edges originating from input $i$ share the same $k$ terms, selected by aggregating $s_{o,i,k}$ over output units. A layer-shared variant uses one support for the entire layer. Per-edge top-$k$ therefore prioritizes functional flexibility, whereas shared-$k$ exposes the regular term dimension required for compact execution. Alternative scoring rules are evaluated in Appendix~\ref{app:selection}.

\paragraph{Neuron and channel sparsification.}
Basis pruning reduces the complexity of individual edges but leaves network width unchanged. SparseKAN therefore removes complete hidden units, or channels in convolutional layers. Removing output $o$ of layer $\ell$ also removes the corresponding input dimension of layer $\ell+1$, so dead features do not persist as unused columns in subsequent layers. The output layer is kept intact. Details of unit scoring and the additional block-structured variant appear in Appendix~\ref{app:structured}.

These two structural axes act on different dimensions of the dominant coefficient tensor. After compression, its first-order cost changes from
\begin{equation}
d_{\ell-1}d_{\ell}K_{\ell}
\quad\longrightarrow\quad
d'_{\ell-1}d'_{\ell}K'_{\ell},
\label{eq:structural_scaling}
\end{equation}
so basis and width reductions compound rather than compete.

\subsection{Physical Compaction and Low-Bit Representation}
\label{sec:compact_quant}

A mask alone does not make a model physically smaller. For structurally compatible supports, SparseKAN therefore slices dead input/output dimensions and gathers the retained shared terms into a coefficient tensor of shape $d'_{\ell}\times d'_{\ell-1}\times K'_{\ell}$. A small \texttt{kept\_term\_idx} table records which original basis functions correspond to the compact term dimension. Thus the resulting model contains smaller tensors rather than full-size tensors populated with zeros. Detailed indexing and the compact forward expression are given in Appendix~\ref{app:compaction}.

The third compression axis reduces the precision of the surviving parameters. SparseKAN applies weight-only, symmetric per-output-channel QAT to both base weights and basis coefficients, using $8$- and $4$-bit operating points. The structural and precision reductions are summarized by the normalized
joint bit-cost
\begin{equation}
\rho_{\mathrm{bit}}
=
\rho_{\mathrm{struct}}\frac{b}{32},
\label{eq:joint_bit_cost}
\end{equation}
where $\rho_{\mathrm{struct}}$ is the final active structural cost and $b$ is the weight precision. Thus basis and neuron sparsification determine
\emph{how much} computation survives, while quantization determines how many bits represent each survivor.

\subsection{Training and Compression Pipeline}
\label{sec:pipeline}

SparseKAN proceeds in four conceptual steps. \textbf{(1) Cost-gated training:} model and gate parameters are optimized jointly using Eq.~\eqref{eq:training_obj}. \textbf{(2) Structural hardening:} continuous gates are binarized and the
requested per-edge/shared-$k$ and neuron/channel constraints are imposed. \textbf{(3) Recovery:} the discrete support is fixed and the surviving parameters are fine-tuned; low-bit models enable QAT during this stage. \textbf{(4) Physical realization:} compatible structured supports are compacted by slicing dead dimensions and gathering retained terms. The resulting operating point is controlled independently by the basis budget $k$, neuron keep ratio $r_{\mathrm n}$, and precision $b$. Implementation variants and the full algorithm are provided in Appendix~\ref{app:full_method}.


\section{Experiments and Analysis}
\label{sec:experiments}

We evaluate SparseKAN around four questions: \textbf{(i)} where does compressible redundancy occur inside a KAN, \textbf{(ii)} do basis and neuron compression provide complementary operating points, \textbf{(iii)} how robust is the surviving model to low-bit representation, and \textbf{(iv)} do these reductions translate into physically smaller and faster implementations? Throughout, we distinguish \emph{active cost}, \emph{physical model size}, and \emph{measured execution cost}. A sparse mask may reduce the first without changing the latter two.

\subsection{Setup and Statistical Protocol}
\label{sec:exp_setup}

Our primary benchmarks are MNIST, CIFAR-10, and CIFAR-100. The main MLP families are EfficientKAN with a B-spline basis and KAGN with a Gram-polynomial basis; convolutional experiments use spatial-kernel KAGN convolutions. Additional KAN families and tabular benchmarks are reported in Appendix~\ref{app:variant_transfer}.

Two complementary campaigns support the results. A three-seed frontier study contains 380 completed runs over seeds $\{42,43,44\}$ and maps the structural and precision operating space. A separate five-seed paired ablation campaign over seeds $42$--$46$ tests comparative claims involving selection, sparsity granularity, gates, and low-bit training. For these comparisons we emphasize paired effect sizes, seed consistency, and confidence intervals. With only five non-zero pairs, the minimum attainable two-sided exact Wilcoxon signed-rank $p$-value is $0.0625$; we therefore do not interpret a small nominal difference as significant when its confidence interval contains zero.

Efficiency is reported using normalized active cost $\rho_{\mathrm{struct}}$, where $1$ denotes dense computation, physical parameter counts measured directly from compact checkpoints, and joint bit-cost $\rho_{\mathrm{bit}}=\rho_{\mathrm{struct}}(b/32)$. Hardware experiments target a ZCU104 using Vitis HLS 2025.2.1 and a 200\,MHz clock target. Hardware methodology is detailed in Sec.~\ref{sec:hardware} and Appendix~\ref{app:fpga_full}.

\subsection{Where Does the Redundancy Lie?}
\label{sec:results_structure}

The structure-discovery stage itself incurs essentially no accuracy penalty: MNIST EfficientKAN changes from $97.05\pm0.28\%$ to $97.07\pm0.24\%$, CIFAR-10 KAGN-conv from $81.73\pm0.62\%$ to $81.41\pm0.58\%$, and wide CIFAR-100 KAGN-conv from $52.72\pm0.45\%$ to $53.24\pm0.52\%$. The near-dense cost under the default schedule is deliberate: a direct regularization sweep moves cost from approximately $0.996$ to $0.52$ while MNIST accuracy changes by less than $0.3$ points. Stage~1 is therefore used as a common adapted checkpoint, with the desired structure imposed explicitly during hardening.

Table~\ref{tab:struct_frontier} makes the two structural axes explicit. Basis reduction is often the gentler first step. MNIST EfficientKAN retains $97.14\pm0.11\%$ with input-shared $k=4$ at cost $0.553$, while KAGN retains only two of four Gram terms at cost $0.600$ and reaches $98.20\pm0.21\%$. The same axis remains useful on harder tasks: shared-$k=3$ gives $80.67\pm0.42\%$ at cost $0.799$ on CIFAR-10 and $53.39\pm0.91\%$ at cost $0.796$ on CIFAR-100.

\begin{table}[!ht]
\centering
\caption{\textbf{Representative structural frontier} (accuracy in \%, three seeds). Basis and neuron budgets expose distinct cost--accuracy trade-offs. The CIFAR-10 Gram model shows the term-axis cliff: shared-$k=2$ is dominated by neuron keep-$0.75$ at similar cost.}
\label{tab:struct_frontier}
\resizebox{\columnwidth}{!}{%
\begin{tabular}{llrr}
\toprule
Model & Operating point & Cost & Accuracy \\
\midrule
MNIST EffKAN & dense & 1.000 & $97.05\pm0.28$ \\
 & shared-$k=4$ & 0.553 & $97.14\pm0.11$ \\
 & neuron keep 0.50 & 0.465 & $95.69\pm0.20$ \\
\midrule
MNIST KAGN & dense & 1.000 & $98.07\pm0.11$ \\
 & shared-$k=2$ & 0.600 & $98.20\pm0.21$ \\
 & neuron keep 0.50 & 0.465 & $97.70\pm0.15$ \\
\midrule
CIFAR-10 conv & dense & 1.000 & $81.73\pm0.62$ \\
 & shared-$k=3$ & 0.799 & $80.67\pm0.42$ \\
 & shared-$k=2$ & 0.600 & $75.32\pm4.47$ \\
 & neuron keep 0.75 & 0.583 & $78.65\pm0.59$ \\
\midrule
CIFAR-100 conv & dense & 1.000 & $52.72\pm0.45$ \\
 & shared-$k=3$ & 0.796 & $53.39\pm0.91$ \\
 & shared-$k=2$ & 0.597 & $52.04\pm0.75$ \\
\bottomrule
\end{tabular}%
}
\end{table}

Width reduction is more architecture-dependent because deleting a channel removes every function carried by that feature, rather than simplifying one function. The CIFAR-10 comparison is especially informative: shared-$k=2$ falls to $75.32\pm4.47\%$ at cost $0.600$, whereas neuron keep-$0.75$ reaches a slightly \emph{lower} cost ($0.583$) with $78.65\pm0.59\%$. Once the already-small four-term Gram basis is compressed too far, another removed term eliminates useful function classes rather than redundancy. This is why SparseKAN keeps basis and width as separate controls instead of assuming one universal sparsity ratio.

\subsection{Which Basis Functions Survive Matters---Sometimes}
\label{sec:selection}

Reducing the basis raises a second question: is the number of surviving functions enough, or does their identity matter? We compare coefficient-based selection with random selection and low-order truncation at identical $k$ and recovery budgets.

\begin{table}[!ht]
\centering
\caption{\textbf{Basis-selection controls.} Coefficient-based selection minus matched low-order truncation and random selection in accuracy points (five-seed campaign; representative rows).}
\label{tab:selection}
\resizebox{0.80\columnwidth}{!}{%
\begin{tabular}{lrrr}
\toprule
Model & $k$ & vs. trunc. & vs. random \\
\midrule
MNIST KAGN & 2 & $+15.25$ & $+0.59$ \\
CIFAR-10 KAGN-conv & 2 & $+8.47$ & $+3.57$ \\
CIFAR-100 KAGN-conv & 2 & $+2.42$ & $-0.08$ \\
MNIST ChebyKAN & 2 & $-0.04$ & $+0.20$ \\
MNIST EfficientKAN & 4 & $+0.13$ & $+0.16$ \\
CIFAR-10 EfficientKAN & 4 & $+0.33$ & $+0.04$ \\
\bottomrule
\end{tabular}%
}
\end{table}

For several Gram-polynomial models, learned selection is critical: low-order truncation trails by $15.25$ points on MNIST KAGN, $8.47$ on CIFAR-10 KAGN-conv, and $2.42$ on CIFAR-100 KAGN-conv. Against random selection, the corresponding MNIST and CIFAR-10 gains are $0.59$ and $3.57$ points. The effect is not universal: ChebyKAN is essentially tied and one MNIST convolutional setting reverses. The supported conclusion is therefore \emph{basis-dependent selection sensitivity}, not a universal polynomial rule. In the tested B-spline models, all selectors stay within $0.33$ points.

Raw gate magnitude is also a poor post-training selector. At identical cost, MNIST KAGN obtains $97.67\%$ with coefficient scoring but only $82.75\%$ with gate scoring; CIFAR-10 KAGN-conv gives $75.22\%$ versus $63.92\%$. Gate saturation therefore makes the learned gate value itself unsuitable as the final importance score.

\subsection{Comparison with Conventional Pruning and Matched MLPs}
\label{sec:baselines}

Table~\ref{tab:baselines_main} summarizes the external controls. At matched cost, basis-level sparsity reaches essentially the same frontier as edge-level $L_0$ sparsity: the gap changes sign across datasets and remains within seed variation. We therefore claim \emph{feasibility rather than accuracy dominance}: moving the decision inside the functional edge preserves the conventional frontier while exposing shared basis supports and basis-specific selection.

A tuned KAN $L_1$+entropy node-pruning baseline is also competitive. SparseKAN's shared-$k$ or neuron operating points match or modestly improve representative cost--accuracy points on CIFAR-10 and CIFAR-100 while using the same framework that later supports basis compaction and quantization. The matched-parameter MLP control provides an important scope condition: KAN wins decisively on CIFAR-10 at the same parameter count, but the MLP wins on MNIST. SparseKAN is therefore useful where the KAN representation first earns its additional functional complexity; it is not an argument for universal KAN superiority.

\begin{table}[!ht]
\centering
\caption{\textbf{Baseline comparisons.} Edge-$L_0$: five paired seeds at matched achieved cost. Node pruning: representative frontier points. Matched MLP: identical dense parameter counts. Accuracy in \%.}
\label{tab:baselines_main}
\resizebox{\columnwidth}{!}{%
\begin{tabular}{llrr}
\toprule
Comparison & Dataset & SparseKAN / KAN & Baseline \\
\midrule
Edge-$L_0$ @ cost 0.554 & MNIST & 97.22 & 97.33 \\
Edge-$L_0$ @ cost 0.600 & CIFAR-10 & 79.07 & 76.56 \\
Edge-$L_0$ @ cost 0.598 & CIFAR-100 & 51.85 & 52.56 \\
\midrule
$L_1$+entropy pruning & CIFAR-10 & 80.67 @ 0.799 & 80.01 @ 0.854 \\
$L_1$+entropy pruning & CIFAR-100 & 53.39 @ 0.796 & 52.36 @ 0.864 \\
\midrule
Matched-param MLP & CIFAR-10 & $90.73\pm0.13$ & $44.39\pm0.58$ \\
Matched-param MLP & MNIST & $97.44\pm0.15$ & $98.22\pm0.06$ \\
\bottomrule
\end{tabular}%
}
\end{table}

\subsection{The Structural Axes Compose and Physically Compact}
\label{sec:results_compaction}

The two structural controls compound predictably. For MNIST EfficientKAN, neuron pruning contributes a cost factor $0.4651$ and top-$k$ basis hardening a factor $0.556$, predicting $0.4651\times0.556=0.2585$; the measured composed cost is $0.2584$. For CIFAR-10 KAGN-conv, $0.2783\times0.602=0.1675$ predicts $0.1676$. The axes therefore modify distinct dimensions of the dominant coefficient tensor, consistent with Eq.~\eqref{eq:structural_scaling}. More importantly, the resulting support can be realized as a genuinely smaller dense model.

\begin{table}[!ht]
\centering
\caption{\textbf{Physical compaction.} Operating points are written sk$k$+n$r$, denoting shared-$k$ basis hardening combined with neuron keep ratio $r$. Reduction is measured from the compact checkpoint rather than inferred from a mask. $\Delta$ is relative to the dense anchor of the same architecture.}
\label{tab:compact}
\resizebox{\columnwidth}{!}{%
\begin{tabular}{llrrr}
\toprule
Dataset & Compact point & Reduction & Acc. & $\Delta$ \\
\midrule
MNIST & EffKAN sk4+n0.5 & $73.0\%$ & 97.98 & $+0.93$ \\
MNIST & KAGN sk2+n0.5 & $70.3\%$ & 97.97 & $-0.10$ \\
CIFAR-10 & Conv sk3+n0.7 & $59.8\%$ & 78.81 & $-2.92$ \\
CIFAR-100 & Conv sk3+n0.7 & $58.4\%$ & 51.45 & $-1.27$ \\
\bottomrule
\end{tabular}%
}
\end{table}

MNIST EfficientKAN removes $73.0\%$ of its parameters with no degradation, and KAGN removes $70.3\%$ for only $0.10$ points. Wide CIFAR-100 removes $58.4\%$ for a $1.27$-point difference. CIFAR-10 is the less redundant case: $59.8\%$ reduction costs $2.92$ points, and more aggressive channel removal degrades rapidly. Extending recovery from 20 to 40 epochs raises one $79.2\%$-reduction point from $67.48\%$ to $71.13\%$, indicating that part of the severe-cut loss is adaptation difficulty.

Physical realization also changes the runtime conclusion. Masked CUDA models remain at $0.95$--$1.03\times$ dense latency because their tensor shapes are unchanged. In contrast, physically compacted convolutional models become faster once the batch is large enough to amortize launch and basis-evaluation overhead (Table~\ref{tab:cuda_main}).

\begin{table}[!ht]
\centering
\caption{\textbf{Physical sparsity translates to CUDA speedup.} Compact/dense latency ratio; values below one are faster. ``Kept'' is the fraction of dense parameters remaining. Masked models stay near $1\times$ and are omitted.}
\label{tab:cuda_main}
\resizebox{\columnwidth}{!}{%
\begin{tabular}{lrrr}
\toprule
Compact point & Params kept & $b=256$ & $b=1024$ \\
\midrule
C10 sk3+n0.7 & $40.2\%$ & 0.79 & 0.76 \\
C10 sk3+n0.5 & $20.8\%$ & 0.54 & 0.54 \\
C100 sk3+n0.7 & $41.6\%$ & 0.75 & 0.76 \\
C100 sk2+n0.5 & $17.1\%$ & 0.53 & 0.51 \\
MNIST EffKAN sk4+n0.5 & $27.0\%$ & 1.03 & 0.82 \\
\bottomrule
\end{tabular}%
}
\end{table}

The result also exposes the remaining systems bottleneck. Parameter count falls faster than latency: the CIFAR-10 sk3+n0.5 model keeps only $20.8\%$ of parameters yet still requires $0.54\times$ dense latency at batch 1024. Basis construction precedes the compressed contraction and therefore remains a target for fused kept-basis kernels. The MLP row shows the complementary limitation: even after $73.0\%$ parameter removal, small dense linear-algebra kernels remain comparatively memory/launch bound.

\subsection{Low Precision: Eight Bits Are Robust, Four Bits Need Adaptation}
\label{sec:quant}

Across the QAT grid, 8-bit weights are effectively free: the largest observed decrease is $0.51$ accuracy points, and several sparse cells slightly improve. Four bits expose a harder regime. MNIST loses only $0.12$--$0.43$ points relative to corresponding FP32 sparse configurations, whereas the evaluated CIFAR-10 settings lose $1.39$--$3.70$ points and CIFAR-100 loses approximately $2.4$--$2.9$ points.

\begin{table}[!ht]
\centering
\caption{\textbf{The 4-bit regime} (accuracy in \%, five seeds). Dense PTQ uses the unpruned model; two-stage and joint configurations use aggressive sparse operating points. Joint-versus-two-stage paired confidence intervals include zero.}
\label{tab:quant4}
\resizebox{\columnwidth}{!}{%
\begin{tabular}{lrrr}
\toprule
Dataset & Dense PTQ-4 & Two-stage-4 & Joint-4 \\
\midrule
MNIST & $97.15\pm0.18$ & $97.14\pm0.08$ & $97.08\pm0.14$ \\
CIFAR-10 & $50.19\pm11.78$ & $75.26\pm4.19$ & $79.72\pm0.30$ \\
CIFAR-100 & $9.04\pm2.43$ & $50.23\pm0.84$ & $51.40\pm0.96$ \\
\bottomrule
\end{tabular}%
}
\end{table}

Dense 4-bit PTQ remains usable on MNIST but collapses on convolutional KANs: $50.19\pm11.78\%$ on CIFAR-10 and $9.04\pm2.43\%$ on CIFAR-100. Quantization-aware adaptation restores useful low-bit operating points. Joint gate/QAT training is notably stable on CIFAR-10 ($79.72\pm0.30\%$ versus $75.26\pm4.19\%$ for the matched two-stage schedule), but its paired confidence interval includes zero and extending the two-stage recovery budget closes most of the mean gap. We therefore claim a robust single-pipeline option, not intrinsic mean-accuracy superiority.

\subsection{Robustness Checks: What Survives Stronger Controls?}
\label{sec:robustness}

The expanded ablation campaign tests the main findings under five-seed paired controls, separating the effects that survive from those that were artifacts of the smaller exploratory study. Table~\ref{tab:robust_main} summarizes the checks most likely to change the interpretation of the main results.

\begin{table}[!ht]
\centering
\caption{\textbf{Robustness audit.} Five-seed paired controls are used for comparative accuracy claims; cost-model and compaction checks use the completed checkpoint set.}
\label{tab:robust_main}
\resizebox{\columnwidth}{!}{%
\begin{tabular}{lll}
\toprule
Check & Evidence & Supported interpretation \\
\midrule
Joint vs. two-stage & $\Delta=-0.06/+4.45/+1.17$ & Parity; lower C10 variance \\
Gate necessity & 5/6 CIs include zero & Gates enable search, not an acc. claim \\
Term vs. edge-$L_0$ & $-0.11/+2.52/-0.71$ pts & Finer granularity at no clear penalty \\
Cost reweighting & Spearman $0.936$--$1.000$ & Frontier ranking is robust \\
Compaction fidelity & top-1 agreement $\geq0.9997$ & Compact model preserves predictions \\
\bottomrule
\end{tabular}%
}
\end{table}

Two points are particularly important. First, the earlier three-seed $+5.69$-point joint-QAS difference and $+4.77$-point training-time-hardening difference are not retained as accuracy claims after five-seed validation. Second, the conclusions are not an artifact of the nominal cost accountant: reweighting 683 completed runs using empirically fitted component costs preserves operating-point order with Spearman correlation $0.936$--$1.000$ across the non-degenerate families. Physical compaction is similarly faithful: across 54 compacted checkpoints, top-1 agreement with the corresponding masked model is at least $0.9997$ (at most three flips per 10,000 examples in the worst case). These checks narrow the paper's claims, but make the remaining ones substantially stronger.

\subsection{FPGA/HLS Realization}
\label{sec:hardware}

We finally instantiate the compact structures in hardware rather than leaving them as software masks. All designs target a ZCU104 (\texttt{xczu7ev-ffvc1156-2-e}) using Vitis HLS 2025.2.1 with a 5\,ns target. Within each family, FPGA part, clock target, AXI interface, weight-storage policy, and HLS pragma policy are fixed across six designs: dense FP32, compact FP32, sparse+int8, sparse+int4, and dense int8/int4 controls.

Hard gates are folded into the exported tensors, so the accelerator contains no gate evaluation or dynamic zero-skipping. Neuron pruning shortens compile-time layer dimensions. Shared-$k$ compression reduces the basis-loop bound, with the retained identities stored in a small \texttt{kept\_term\_idx} table. The accelerator therefore executes a smaller dense problem rather than a dense problem containing zeros. EfficientKAN and GRAM exceed on-chip memory and stream weights from DDR; KAGN-conv is small enough to store its weights in on-chip ROM. All designs meet timing, with achieved clocks between approximately $4.40$ and $4.71$\,ns.

\begin{table}[!ht]
\centering
\caption{\textbf{FPGA/HLS inference ladder.} MLP latency is from RTL co-simulation; KAGN-conv$^{\dagger}$ uses the HLS synthesis worst-case cycle estimate. Accuracy is evaluated on the exported MNIST hardware checkpoints.}
\label{tab:hw_main}
\resizebox{\columnwidth}{!}{%
\begin{tabular}{llrrr}
\toprule
Family & Design & Acc. & Latency & Speedup \\
\midrule
EffKAN & FP32 & 97.48 & 2.872 ms & $1.00\times$ \\
 & Sparse & 98.00 & 0.872 ms & $3.29\times$ \\
 & Sparse+int8 & 97.96 & 0.195 ms & $14.69\times$ \\
 & Sparse+int4 & 97.92 & 0.122 ms & $23.63\times$ \\
 & Dense int4 & 96.83 & 0.366 ms & $7.85\times$ \\
\midrule
GRAM & FP32 & 98.14 & 1.898 ms & $1.00\times$ \\
 & Sparse & 98.22 & 0.659 ms & $2.88\times$ \\
 & Sparse+int8 & 98.11 & 0.151 ms & $12.55\times$ \\
 & Sparse+int4 & 98.22 & 0.104 ms & $18.21\times$ \\
 & Dense int4 & 98.06 & 0.262 ms & $7.25\times$ \\
\midrule
KAGN-conv & FP32 & 98.45 & 86.677 ms & $1.00\times$ \\
 & Sparse+int4 & 98.36 & 57.219 ms & $1.51\times^{\dagger}$ \\
\bottomrule
\end{tabular}%
}
\end{table}

\paragraph{MLPs: compression becomes latency and bandwidth reduction.}
EfficientKAN improves from $2.872$\,ms in FP32 to $0.122$\,ms under sparse int4, a $23.63\times$ reduction; GRAM reaches $18.21\times$. Dense int4 controls reach only $7.85\times$ and $7.25\times$, respectively, isolating the contribution of structural compaction. At int8, the measured stagewise factors compose to rounding ($3.29\times4.46\approx14.69$ for EfficientKAN and $2.88\times4.36\approx12.55$ for GRAM). Because these MLPs stream weights from DDR, the same transformation reduces per-image weight traffic from approximately $8.45$ to $0.27$\,MB for EfficientKAN and $4.70$ to $0.16$\,MB for GRAM. The low-bit MLP designs sometimes consume \emph{more} LUT/DSP resources than FP32 because the HLS implementation trades parallel arithmetic for latency; their hardware result is therefore latency/bandwidth reduction, not area reduction.

\paragraph{KAGN-conv: compression becomes resource headroom.}
The convolutional accelerator exposes a different bottleneck. Its one-MAC-per-cycle engine remains dominated by spatial basis construction and convolution, so sparse int4 improves serial latency by only $1.51\times$. Device occupancy, however, falls sharply.

\begin{table}[!ht]
\centering
\caption{\textbf{KAGN-conv post-synthesis utilization.} Sparse+int4 reduces DSP occupancy from $96.2\%$ to $29.1\%$ with only $-0.09$ points of accuracy change.}
\label{tab:hw_resources}
\resizebox{0.9\columnwidth}{!}{%
\begin{tabular}{lrrrr}
\toprule
Design & DSP & LUT & FF & BRAM \\
\midrule
FP32 & $96.2\%$ & $71.0\%$ & $50.6\%$ & $75.8\%$ \\
Sparse & $77.7\%$ & $60.2\%$ & $42.0\%$ & $68.1\%$ \\
Sparse+int8 & $34.8\%$ & $35.0\%$ & $17.0\%$ & $47.1\%$ \\
Sparse+int4 & $29.1\%$ & $34.8\%$ & $16.6\%$ & $43.9\%$ \\
Dense int4 & $28.4\%$ & $36.3\%$ & $17.1\%$ & $45.4\%$ \\
\bottomrule
\end{tabular}%
}
\end{table}

Sparse+int4 reduces DSP count from 1,663 to 502, LUT utilization from $71.0\%$ to $34.8\%$, and BRAM from $75.8\%$ to $43.9\%$, while accuracy changes from $98.45\%$ to $98.36\%$. Dense int4 reaches similar area but requires $16.5$M rather than $12.7$M cycles per image, showing that quantization supplies most of the arithmetic-area reduction while structural sparsity additionally shortens the executed contraction. The result is primarily \emph{capacity headroom}: the compressed engine leaves room for parallel replication or larger models, although such replication is not implemented here.

\paragraph{Verification.}
All 18 designs are checked against NumPy references generated from the exported artifacts. MLP designs pass RTL co-simulation; integer implementations using the same arithmetic are bit-exact where applicable, while FP32 differences are limited to small floating-point reassociation errors. Full-image RTL co-simulation is impractical for the 12--19M-cycle KAGN-conv models, so their latency is synthesis-estimated and each design is additionally checked on 64 C-simulation images. All six convolutional designs obtain 64/64 argmax agreement.

\subsection{Discussion and Limitations}
\label{sec:results_discussion}

Three conclusions emerge from the study. First, KAN redundancy is \emph{multi-axis and basis-dependent}: basis functions, neurons/channels, and precision expose distinct operating points, and the importance of individual terms depends on the basis family and task. Second, sparsity improves deployment only when it is physically realized; masks leave dense tensor shapes unchanged, whereas structured hardening and compaction reduce storage, CUDA latency, and hardware loop bounds. Third, precision interacts with task difficulty: 8-bit QAT is broadly benign, while 4-bit convolutional KANs require quantization-aware recovery.

The robustness audit also limits the claims appropriately. Basis-level sparsity does not consistently outperform matched-cost edge-$L_0$ pruning, joint QAS does not establish a mean-accuracy advantage over a sufficiently recovered two-stage pipeline, and gate training is not itself an accuracy contribution. SparseKAN's value is instead the common interface it provides for discovering, controlling, and physically realizing functional, structural, and precision compression across KAN families.

\section{Conclusion}
\label{sec:conclusion}

SparseKAN addresses a compression dimension specific to KANs: redundancy can occur not only across edges and neurons, but also within the basis expansion of each functional edge. The framework exposes basis functions, neurons/channels, and precision as separately controllable axes, then converts the selected structure into compact tensors through hardening, recovery, and physical compaction. Experiments across multiple KAN families show that basis and width reductions compose predictably, while basis selection can be critical in Gram-polynomial models. Compaction removes up to $73.0\%$ of parameters without loss on MNIST and reduces large-batch CUDA latency to $0.51\times$ dense execution. On a ZCU104, sparse low-bit MLPs achieve up to $23.63\times$ lower latency, while KAGN-conv reduces DSP occupancy from $96.2\%$ to $29.1\%$ with negligible accuracy change. These results establish SparseKAN as a basis-aware route from learned KAN redundancy to compact, low-bit software and hardware implementations.

\bibliography{aaai2027} 

\clearpage
\appendix

\section{Additional SparseKAN Method Details}
\label{app:full_method}

This appendix provides the training, hardening, compaction, and quantization
details omitted from Sec.~\ref{sec:method} for space.

\subsection{Gate Relaxations and Hardening}
\label{app:gates}

The default SparseKAN gate is a sigmoid-relaxed learnable scalar
\begin{equation}
g=\sigma(\alpha),
\qquad g\in(0,1),
\label{eq:app_sigmoid}
\end{equation}
where $\alpha$ is the gate logit. The implementation additionally supports
hard-concrete gates~\citep{louizos2018l0} and a binary
Gumbel straight-through relaxation. These choices modify the stochastic
training-time surrogate but not the downstream compression interface.
At hardening, each soft gate is mapped to a deterministic binary decision
using threshold $\tau$; the corresponding logits are then saturated to large
positive or negative values so that the structure remains fixed during
recovery.

For convolutional KANs, the same base/branch/term hierarchy is used over
spatial coefficient kernels. A term decision applies to the complete spatial
kernel associated with that basis function rather than independently to each
kernel position.

\subsection{Cost Scheduling and Direct Budget Control}
\label{app:cost}

SparseKAN begins with a warm-up interval in which no sparsity pressure is
applied. Each cost coefficient is then increased linearly toward its target:
\begin{equation}
\lambda_q(e)
=
\lambda_q^{\max}
\begin{cases}
0, & e\le E_{\mathrm w},\\[1mm]
\min\!\left(
1,\frac{e-E_{\mathrm w}}{E_{\mathrm r}}
\right), & e>E_{\mathrm w},
\end{cases}
\qquad
q\in\{\mathrm b,\mathrm r,\mathrm t\},
\label{eq:app_lambda_schedule}
\end{equation}
where $E_{\mathrm w}$ and $E_{\mathrm r}$ are the warm-up and ramp durations.

The framework also supports direct cost-budget control. Let
$\rho_{\mathrm{soft}}$ denote the normalized soft active cost evaluated
without the fixed $\lambda_q$ multipliers. Given target $\theta$, we optimize
\begin{equation}
\mathcal{R}_{\mathrm{budget}}
=
\mu[\rho_{\mathrm{soft}}-\theta]_+,
\label{eq:app_budget}
\end{equation}
where $[z]_+=\max(0,z)$ and the dual variable is updated by projected ascent,
\begin{equation}
\mu
\leftarrow
\max\!\left(
0,\,
\mu+\eta_\mu
(\rho_{\mathrm{soft}}-\theta)
\right).
\label{eq:app_dual}
\end{equation}
This mode is used when a desired computational budget should be specified
directly rather than through a manually chosen regularization coefficient.

Under the default schedule, Stage~1 is intentionally conservative and often
finishes close to the dense soft-cost point. It should therefore be understood
as a \emph{structure-discovery and adaptation stage}, not as the final
compressed operating point. Explicit compression is imposed by the hardening
operations below.

\subsection{Basis Selection Rules}
\label{app:selection}

\paragraph{Per-edge top-$k$.}
Given the default score from Eq.~\eqref{eq:term_score},
\begin{equation}
\mathcal{S}^{(k)}_{o,i}
=
\operatorname{TopK}_{k}
\{s_{o,i,1},\ldots,s_{o,i,K}\},
\label{eq:app_edge_topk}
\end{equation}
and
\begin{equation}
\hat g^{\mathrm t}_{o,i,k}
=
\mathbb{I}
[k\in\mathcal{S}^{(k)}_{o,i}]
\,
\mathbb{I}
[g^{\mathrm r}_{o,i}>\tau].
\label{eq:app_edge_mask}
\end{equation}

\paragraph{Input-shared top-$k$.}
All outgoing edges associated with input $i$ share one support:
\begin{equation}
\bar{s}_{i,k}
=
\sum_o s_{o,i,k},
\qquad
\mathcal{S}^{(k)}_i
=
\operatorname{TopK}_{k}
\{\bar{s}_{i,1},\ldots,\bar{s}_{i,K}\}.
\label{eq:app_shared_input}
\end{equation}

\paragraph{Layer-shared top-$k$.}
The most regular variant uses
\begin{equation}
\bar{s}_k
=
\sum_i\sum_o s_{o,i,k},
\qquad
\mathcal{S}^{(k)}
=
\operatorname{TopK}_{k}
\{\bar{s}_1,\ldots,\bar{s}_K\}.
\label{eq:app_shared_layer}
\end{equation}

\paragraph{Alternative scores.}
For controlled ablations the implementation also provides:
\begin{align}
s^{\mathrm{gate}}_{o,i,k}
&=g^{\mathrm t}_{o,i,k},
\\
s^{\mathrm{occ}}_{o,i,k}
&=\rho_{i,k}|c_{o,i,k}|g^{\mathrm t}_{o,i,k},
\\
s^{\mathrm{quant}}_{o,i,k}
&=\rho_{i,k}|Q_b(c_{o,i,k})|g^{\mathrm t}_{o,i,k},
\end{align}
where $\rho_{i,k}$ denotes empirical basis occupancy and $Q_b$ is the
$b$-bit fake-quantization operator. These scores are experimental
alternatives rather than prerequisites of the core framework.

\subsection{Structured Neuron, Channel, and Block Pruning}
\label{app:structured}

SparseKAN supports neuron/channel sparsification in both the in-training
structured ablation and the final compaction path. These use related but
slightly different rankings.

During generic structured hardening, an output unit is scored from the
row-wise activity of its base, branch, and term gates. After shared-$k$
hardening, however, many gate values can become saturated and nearly
indistinguishable across units. For final physical compaction we therefore
rank output $o$ using the magnitude of its surviving computation:
\begin{equation}
u_o
=
\sum_i
g^{\mathrm b}_{o,i}|w^{\mathrm b}_{o,i}|
+
\sum_{i,k}
g^{\mathrm r}_{o,i}
g^{\mathrm t}_{o,i,k}
|c_{o,i,k}|.
\label{eq:app_unit_score}
\end{equation}
For convolutional KANs, the coefficient magnitude is additionally summed over
the spatial kernel dimensions.

Given keep ratio $r_{\mathrm n}$,
\begin{equation}
d'_\ell
=
\max\!\left(
1,\operatorname{round}(r_{\mathrm n}d_\ell)
\right)
\label{eq:app_neuron_keep}
\end{equation}
highest-scoring hidden units are retained. Removing output unit $o$ from
layer $\ell$ simultaneously removes its corresponding input dimension from
layer $\ell+1$. The final network output is protected.

The implementation also supports block-structured pruning, in which groups
of neighboring coefficient connections are hardened together. Block
sparsity is used as a regularity ablation and is not required by the main
SparseKAN pipeline.

\subsection{Physical Compaction}
\label{app:compaction}

Let $\mathcal I_\ell$ and $\mathcal O_\ell$ denote the surviving input and
output indices of layer $\ell$. The compact base matrix is
\begin{equation}
\widetilde{\mathbf W}^{\mathrm b,\ell}
=
\mathbf W^{\mathrm b,\ell}
[\mathcal O_\ell,\mathcal I_\ell].
\label{eq:app_compact_base}
\end{equation}
The same index sets slice the corresponding coefficient, gate, and
normalization tensors.

For input-shared basis sparsity, let
$\mathcal K_{\ell,i}^{(k)}$ denote the retained basis indices associated with
input $i$. The compact coefficient tensor is formed as
\begin{equation}
\widehat{\mathbf C}^{\ell}_{:,i,:}
=
\mathbf C^\ell_{
\mathcal O_\ell,\,
i,\,
\mathcal K_{\ell,i}^{(k)}
},
\label{eq:app_compact_terms}
\end{equation}
giving shape
$d'_\ell\times d'_{\ell-1}\times k$.
The original basis identities are stored in
\begin{equation}
\mathbf R^\ell
\in
\{1,\ldots,K_\ell\}^{d'_{\ell-1}\times k},
\label{eq:app_kept_idx}
\end{equation}
implemented by \texttt{kept\_term\_idx}. Layer-shared sparsity is the special
case in which every row of $\mathbf R^\ell$ is identical.

The compact forward computation is
\begin{equation}
\widetilde y_o
=
\sum_{i=1}^{d'_{\ell-1}}
\widetilde w^{\mathrm b}_{o,i}a(x_i)
+
\sum_{i=1}^{d'_{\ell-1}}
\sum_{j=1}^{k}
\widehat c_{o,i,j}
\psi_{R^\ell_{i,j}}(x_i).
\label{eq:app_compact_forward}
\end{equation}
Consequently, the dominant coefficient storage changes from
$d_{\ell-1}d_\ell K_\ell$ to approximately
$d'_{\ell-1}d'_\ell k$, plus the much smaller retained-index table.
Arbitrary per-edge top-$k$ does not share a common basis dimension and is
therefore not physically gathered by this regular compact representation.

After aggressive structural cuts, a recovery or ``healing'' stage may be
used so that the surviving parameters adapt to the compact architecture.

\subsection{Quantization-Aware Training}
\label{app:quantization}

SparseKAN uses weight-only symmetric fake quantization of the base weights and
basis coefficients. For target precision $b<32$,
\begin{equation}
q_{\max}=2^{b-1}-1.
\end{equation}
For output channel $o$, the per-channel scale is
\begin{equation}
s_o
=
\frac{
\max_{\mathbf w\in\mathbf W_o}|\mathbf w|
}{
q_{\max}
},
\label{eq:app_quant_scale}
\end{equation}
and
\begin{equation}
Q_b(\mathbf W_o)
=
s_o\,
\operatorname{clip}
\left(
\operatorname{round}
\left(
\frac{\mathbf W_o}{s_o}
\right),
-q_{\max},q_{\max}
\right).
\label{eq:app_fake_quant}
\end{equation}
The underlying FP32 parameter is retained during optimization using the
straight-through construction
\begin{equation}
\widetilde{\mathbf W}
=
\mathbf W+
\operatorname{sg}
\left(
Q_b(\mathbf W)-\mathbf W
\right),
\label{eq:app_quant_ste}
\end{equation}
so the forward pass observes fake-quantized weights while gradients propagate
through the full-precision parameter. Activations remain unquantized in the
software SparseKAN experiments.

\subsection{Quantization-Aware Sparsification Extensions}
\label{app:qas}

The default pipeline first establishes structural sparsity and then enables
QAT during recovery. We additionally implement two tighter couplings between
structure and precision.

\paragraph{Joint gate learning and QAT.}
Fake quantization may be activated during Stage~1 so that gates are optimized
under quantized forward semantics rather than being learned entirely in
FP32. We evaluate this as an alternative training schedule rather than a
required part of SparseKAN.

\paragraph{Learned layer-wise precision.}
Each sparse layer can instead carry a continuous bit variable
$\beta_\ell\in[b_{\min},b_{\max}]$, optimized jointly with the network using
\begin{equation}
\mathcal R_{\mathrm{bits}}
=
\frac{
\sum_\ell
C_\ell^{\mathrm{soft}}\beta_\ell
}{
C_{\mathrm{dense}}\,32
}.
\label{eq:app_learned_bits}
\end{equation}
After the structural training stage, $\beta_\ell$ is rounded to an integer
and frozen. This provides an optional mixed-precision search mechanism. The
main-paper results, however, use fixed 8- and 4-bit settings unless stated
otherwise.

\subsection{Complete Training Procedure}
\label{app:algorithm}

\begin{algorithm}[t]
\caption{SparseKAN Training and Compression}
\label{alg:sparsekan}
\begin{algorithmic}[1]
\Require Dataset $\mathcal D$, KAN $f_\theta$, term budget $k$,
neuron keep ratio $r_{\mathrm n}$, bit-width $b$,
hardening threshold $\tau$
\State Attach hierarchical gates
$\{g^{\mathrm b},g^{\mathrm r},g^{\mathrm t}\}$ to each sparse KAN layer
\For{$e=1,\ldots,E$}
    \State Update cost coefficients using
    Eq.~\eqref{eq:app_lambda_schedule}
    \State Optimize model and gates using Eq.~\eqref{eq:training_obj}
\EndFor
\State Harden continuous gates at threshold $\tau$
\If{basis sparsification is enabled}
    \State Apply per-edge top-$k$, input-shared $k$, or layer-shared $k$
\EndIf
\If{neuron/channel compression is enabled}
    \State Retain the selected hidden dimensions and propagate removals
          across adjacent layers
\EndIf
\If{$b<32$}
    \State Enable per-output-channel $b$-bit fake quantization
\EndIf
\State Fine-tune the surviving model with structural support fixed
\If{the selected support is compactable}
    \State Slice dead dimensions and gather shared retained terms
\EndIf
\State \Return compressed SparseKAN
\end{algorithmic}
\end{algorithm}
\section{Additional Experimental Results}
\label{app:results}

This section provides the complete evidence supporting Sec.~\ref{sec:experiments}. We separate the three-seed frontier study, which maps available operating points, from the five-seed paired ablation campaign used to evaluate comparative claims. We further distinguish analytical active cost, compact parameter count, measured CUDA latency, and FPGA/HLS results.

\subsection{Experimental Protocol}
\label{app:exp_details}

The primary frontier consists of 380 completed runs using seeds $\{42,43,44\}$. Unless stated otherwise, frontier results are mean $\pm$ sample standard deviation over these seeds.

The follow-up ablation campaign uses five paired seeds $\{42,43,44,45,46\}$ for comparisons involving selection controls, edge-level baselines, gate necessity, PTQ, and joint versus staged low-bit training. Paired seeds are shared between arms wherever possible. At $n=5$, an exact two-sided Wilcoxon signed-rank test cannot attain a $p$-value below $0.0625$ for five non-zero paired differences. We therefore report effect size, sign consistency, confidence intervals, and seed spread rather than using $p<0.05$ as the sole criterion.

The normalized active-cost ratio is denoted $\rho_{\mathrm{struct}}$. For bit-width $b$, the joint bit-cost is $\rho_{\mathrm{bit}}=\rho_{\mathrm{struct}}(b/32)$. Physical reductions are computed directly from compact checkpoints rather than inferred from binary masks.

\subsection{Five-Seed Claim Audit}
\label{app:claim_audit}

Because several exploratory three-seed differences were large, we explicitly re-tested the corresponding comparative claims with five paired seeds. Table~\ref{tab:app_claim_audit} records the resulting evidence hierarchy. This table is intended to make clear which conclusions are used as positive claims in the paper and which are retained only as parity, stability, or scope observations.

\begin{table*}[!ht]
\centering
\caption{\textbf{Five-seed audit of the main comparative claims.} $\Delta$ is the paired SparseKAN/joint-minus-control difference in accuracy points unless stated otherwise. ``Parity'' means the observed difference is within seed variation or its confidence interval includes zero.}
\label{tab:app_claim_audit}
\resizebox{\textwidth}{!}{%
\begin{tabular}{llll}
\toprule
Question & Result & Verdict & Main-paper use \\
\midrule
Joint QAS vs. two-stage & MNIST $-0.06$; C10 $+4.45\pm4.17$; C100 $+1.17\pm1.36$ & Not separable & Stability/parity only \\
Joint vs. extended two-stage & $-0.12$ / $+2.66$ / $-0.27$ & Parity & No accuracy-win claim \\
Gate-trained vs. gate-free & 5/6 arms CI includes 0 & Mostly null & Gates justified by functionality \\
Learned vs. truncation & KAGN $+15.25$; C10 conv $+8.47$; C100 conv $+2.42$ & Basis-dependent positive & Main selection claim \\
Learned vs. random & C10 conv $+3.57$; MNIST KAGN $+0.59$; splines $\approx0$ & Basis-dependent & Supporting selection claim \\
Term-level vs. edge-$L_0$ & $-0.11$ / $+2.52$ / $-0.71$ & Matched-cost parity & Feasibility, not dominance \\
Matched-param KAN vs. MLP & C10: $90.73$ vs. $44.39$; MNIST: $97.44$ vs. $98.22$ & Dataset-dependent & Scope condition \\
Dense 4-bit PTQ & C10 $50.19\pm11.78$; C100 $9.04\pm2.43$ & Conv collapse & Motivates QAT \\
Cost-model reweighting & Spearman $\rho=0.936$--$1.000$ over non-degenerate families & Rank-robust & Cost conclusions preserved \\
Physical compaction & top-1 agreement $\geq0.9997$ over 54 checkpoints & Near-equivalent & Supports physical realization \\
\bottomrule
\end{tabular}%
}
\end{table*}

The audit changes the interpretation of two exploratory observations. First, the original $+5.69$-point joint-QAS headline is not retained: a matched five-seed comparison is not separable, and extending the recovery budget of the staged baseline closes most of the mean gap. Second, the original $+4.77$-point training-time-versus-post-hoc hardening observation also falls within seed variation. Conversely, the basis-selection result survives the expanded controls and becomes stronger mechanistically because the same family dependence appears against both low-order truncation and random selection. The engineering conclusions are also checked independently: reweighting the cost model preserves the frontier ordering, and physical compaction retains at least $0.9997$ top-1 agreement with the masked source model across all 54 checked checkpoints.

\subsection{Dense Anchors}
\label{app:dense_anchors}

\begin{table*}[!ht]
\centering
\caption{\textbf{Dense and Stage-1 cost-gated anchors.} Accuracy in \%, three seeds. Stage~1 uses the conservative default schedule; explicit compression is imposed during subsequent hardening.}
\label{tab:app_anchors}
\resizebox{0.75\textwidth}{!}{%
\begin{tabular}{llrrrr}
\toprule
Dataset & Variant & Params & Dense acc. & Sparse-trained acc. & Cost \\
\midrule
MNIST & EfficientKAN & 4{,}460{,}288 & $97.05\pm0.28$ & $97.07\pm0.24$ & 0.9959 \\
MNIST & KAGN & 2{,}583{,}072 & $98.07\pm0.11$ & $98.18\pm0.10$ & 1.0000 \\
MNIST & KAGN-conv & 33{,}120 & $97.59\pm0.25$ & $97.75\pm0.15$ & 0.9969 \\
MNIST & ChebyKAN & 2{,}582{,}272 & $97.59\pm0.17$ & $97.41\pm0.10$ & 0.9534 \\
MNIST & FastKAN & 4{,}463{,}018 & $97.49\pm0.21$ & $97.51\pm0.11$ & 0.9964 \\
MNIST & PyKAN & 4{,}695{,}040 & $98.20\pm0.26$ & $98.28\pm0.14$ & 0.9997 \\
MNIST & ReLUKAN & 4{,}460{,}288 & $96.71\pm0.21$ & $96.89\pm0.17$ & 0.9961 \\
MNIST & WavKAN & 4{,}462{,}648 & $96.78\pm0.35$ & $96.83\pm0.41$ & 0.9982 \\
\midrule
CIFAR-10 & EfficientKAN & 16{,}251{,}584 & $89.96\pm0.38$ & $90.30\pm0.35$ & 0.9995 \\
CIFAR-10 & KAGN-conv & 592{,}932 & $81.73\pm0.62$ & $81.41\pm0.58$ & 0.9960 \\
CIFAR-100 & KAGN-conv (narrow) & 719{,}832 & $49.04\pm0.71$ & $49.24\pm0.69$ & 0.9751 \\
CIFAR-100 & KAGN-conv (wide) & 2{,}586{,}328 & $52.72\pm0.45$ & $53.24\pm0.52$ & 0.9683 \\
\bottomrule
\end{tabular}%
}
\end{table*}

\subsection{Variant Transfer}
\label{app:variant_transfer}

\begin{table}[!ht]
\centering
\caption{\textbf{Transfer of post-hoc term sparsification across MNIST KAN families} (accuracy in \%, three seeds). The result demonstrates substantial family dependence rather than a universal polynomial-versus-spline rule.}
\label{tab:app_transfer}
\resizebox{\columnwidth}{!}{%
\begin{tabular}{lrrr}
\toprule
Variant & Dense & keep 0.5 & keep 0.05 \\
\midrule
EfficientKAN & $97.05\pm0.28$ & $97.62\pm0.17$ & $97.48\pm0.08$ \\
KAGN & $98.07\pm0.11$ & $97.36\pm0.32$ & $86.79\pm5.51$ \\
ChebyKAN & $97.59\pm0.17$ & $97.90\pm0.06$ & $97.83\pm0.08$ \\
FastKAN & $97.49\pm0.21$ & $98.00\pm0.16$ & $97.63\pm0.13$ \\
PyKAN & $98.20\pm0.26$ & $98.32\pm0.09$ & $98.27\pm0.10$ \\
ReLUKAN & $96.71\pm0.21$ & $97.38\pm0.06$ & $96.82\pm0.27$ \\
WavKAN & $96.78\pm0.35$ & $97.70\pm0.09$ & $97.16\pm0.13$ \\
KAGN-conv & $97.59\pm0.25$ & $85.77\pm0.08$ & $83.68\pm0.79$ \\
\bottomrule
\end{tabular}%
}
\end{table}

\subsection{Selection Controls}
\label{app:selection_results}

\begin{table*}[!ht]
\centering
\caption{\textbf{Coefficient selection versus low-order truncation and random selection} (five-seed campaign). Large truncation failures occur in several Gram-polynomial models, while the tested B-spline models are largely insensitive to support identity.}
\label{tab:app_selection}
\resizebox{0.75\textwidth}{!}{%
\begin{tabular}{llrrrr}
\toprule
Model & Basis & $k$ & Coeff. acc. & $\Delta$ vs. trunc. & $\Delta$ vs. random \\
\midrule
MNIST KAGN & Gram polynomial & 2 & $97.73\pm0.17$ & $+15.25$ & $+0.59$ \\
CIFAR-10 KAGN-conv & Gram polynomial & 2 & 79.07 & $+8.47$ & $+3.57$ \\
CIFAR-100 KAGN-conv & Gram polynomial & 2 & $40.82\pm1.00$ & $+2.42$ & $-0.08$ \\
MNIST ChebyKAN & Chebyshev & 2 & 97.9 & $-0.04$ & $+0.20$ \\
MNIST KAGN-conv & Gram polynomial & 2 & 85.9 & $-1.66$ & $+1.74$ \\
MNIST EfficientKAN & B-spline & 2--4 & 97.2--97.6 & $\leq0.13$ & $\leq0.16$ \\
CIFAR-10 EfficientKAN & B-spline & 2--4 & $\approx91.6$ & $\leq0.33$ & $+0.04$ \\
\bottomrule
\end{tabular}%
}
\end{table*}

For MNIST KAGN, coefficient selection exceeds random selection on all five paired seeds; the reported paired interval is $[+0.41,+0.77]$ points. For CIFAR-100 KAGN-conv, the coefficient-versus-random effect is effectively zero. Thus the data supports basis-dependent selection sensitivity, not a universal selector advantage.

The post-hoc scoring sweep further shows that raw gate value is a poor criterion after structured gate saturation. MNIST KAGN falls from $97.67\%$ under coefficient scoring to $82.75\%$ under gate scoring at the same cost, and CIFAR-10 KAGN-conv falls from $75.22\%$ to $63.92\%$. A quantization-aware coefficient score is also worse than ordinary coefficient magnitude in seven of nine evaluated cells and is not used as the default.

\subsection{Matched-Cost Edge-Level Baseline}
\label{app:edge_baseline}

The original edge-level control did not produce non-trivial sparsity and was therefore re-specified using the direct Lagrangian budget mode. The replacement baseline gates edges across all evaluated runs, with budget points chosen to bracket the SparseKAN operating point. The edge frontier is interpolated to the exact SparseKAN cost.

\begin{table}[!ht]
\centering
\caption{\textbf{Basis-function sparsity versus matched-cost edge-$L_0$} (accuracy in \%, five paired seeds). The gaps change sign by dataset and remain within the observed seed spread.}
\label{tab:app_edge}
\resizebox{\columnwidth}{!}{%
\begin{tabular}{lrrr}
\toprule
Family @ cost & SparseKAN & Edge-$L_0$ & Gap \\
\midrule
MNIST EffKAN @ 0.554 & 97.22 & 97.33 & $-0.11$ \\
CIFAR-10 conv @ 0.600 & 79.07 & 76.56 & $+2.52$ \\
CIFAR-100 conv @ 0.598 & 51.85 & 52.56 & $-0.71$ \\
\bottomrule
\end{tabular}%
}
\end{table}

\subsection{KAN Node-Pruning Baseline}
\label{app:node_baseline}

The KAN $L_1$+entropy baseline is trained with three penalty strengths and evaluated over eight node-retention levels, followed by healing. Representative CIFAR-10 points are $80.82\%$ at cost $0.924$, $80.01\%$ at $0.854$, $76.23\%$ at $0.708$, and $64.79\%$ at $0.427$. SparseKAN reaches $80.67\pm0.42\%$ at $0.799$ using shared-$k=3$ and $78.65\pm0.59\%$ at $0.583$ using neuron keep-$0.75$.

For CIFAR-100, representative baseline points are $53.57\%$ at $0.931$, $52.36\%$ at $0.864$, $48.06\%$ at $0.731$, and $34.12\%$ at $0.478$. SparseKAN shared-$k=3$ gives $53.39\pm0.91\%$ at $0.796$. The baseline frontiers are similar across penalty strengths, so these results are used to establish competitiveness rather than a large accuracy advantage.

\subsection{Gate-Training Ablation}
\label{app:gate_ablation}

Gate-trained and gate-free versions use the same downstream hardening procedure. Five of six evaluated arms are not separable at five seeds: representative gated-minus-ungated effects are $+0.13\pm0.16$ points on MNIST FP32, $+0.03\pm0.12$ on MNIST 4-bit, and $+0.16\pm0.30$ on CIFAR-10 FP32.

The largest mean effect appears on the CIFAR-100 4-bit arm (approximately $+1.15$ points; reported interval $[+0.03,+2.27]$), although the sign reverses for one seed. We therefore do not claim that differentiable gate training is necessary for ordinary coefficient hardening. Its value is the common optimization interface it provides for structure discovery, cost budgets, and optional joint structural/precision training.

\subsection{Matched-Parameter MLP Control}
\label{app:mlp_baseline}

On CIFAR-10, the parameter-matched MLP and KAN each begin with 16{,}251{,}584 parameters. The MLP reaches $40.51\pm1.22\%$ when dense and $44.39\pm0.58\%$ in its best evaluated compressed configuration, whereas the corresponding KAN pipeline reaches $90.73\pm0.13\%$. The 4-bit MLP arm reaches $23.75\pm3.56\%$.

The result reverses on MNIST: at the matched 4{,}460{,}288-parameter scale, the MLP reaches $98.22\pm0.06\%$ versus $97.44\pm0.15\%$ for the KAN. This experiment therefore scopes the deployment motivation: KAN compression is useful where the KAN representation first provides a task-level benefit.

\subsection{Full Post-Hoc Selection Frontier}
\label{app:posthoc_frontier}

\begin{table*}[!ht]
\centering
\caption{\textbf{Global keep-ratio and per-edge top-$k$ at approximately matched cost} (accuracy in \%, three seeds). Global importance does not consistently dominate per-edge top-$k$; the more reproducible observation is the failure of raw gate scoring after hardening.}
\label{tab:app_topk}
\resizebox{0.85\textwidth}{!}{%
\begin{tabular}{lrrrrrr}
\toprule
Family & Cost (global/top-$k$) & $k$ & Global & Top-$k$ coeff. & Top-$k$ gate & Top-$k$ occ. \\
\midrule
MNIST EffKAN & 0.553/0.553 & 4 & $97.62\pm0.17$ & $97.61\pm0.05$ & $96.14\pm1.49$ & $97.67\pm0.13$ \\
MNIST KAGN & 0.600/0.600 & 2 & $97.36\pm0.32$ & $97.67\pm0.21$ & $82.75\pm0.21$ & $90.64\pm5.12$ \\
MNIST KAGN-conv & 0.599/0.600 & 2 & $85.77\pm0.08$ & $87.12\pm1.32$ & $87.23\pm0.18$ & $87.10\pm1.30$ \\
MNIST ChebyKAN & 0.576/0.584 & 2 & $97.90\pm0.06$ & $97.89\pm0.13$ & $97.75\pm0.10$ & $97.92\pm0.06$ \\
MNIST FastKAN & 0.553/0.554 & 4 & $98.00\pm0.16$ & $97.89\pm0.07$ & $96.42\pm0.49$ & $97.93\pm0.09$ \\
MNIST PyKAN & 0.555/0.555 & 4 & $98.32\pm0.09$ & $98.11\pm0.17$ & $98.18\pm0.10$ & $98.20\pm0.10$ \\
MNIST ReLUKAN & 0.553/0.554 & 4 & $97.38\pm0.06$ & $97.18\pm0.17$ & $93.73\pm1.00$ & $97.26\pm0.08$ \\
MNIST WavKAN & 0.554/0.554 & 4 & $97.70\pm0.09$ & $97.59\pm0.12$ & $96.15\pm0.64$ & $97.62\pm0.11$ \\
C10 EffKAN & 0.555/0.555 & 4 & $91.53\pm0.14$ & $91.66\pm0.25$ & $91.52\pm0.08$ & $91.52\pm0.15$ \\
C10 KAGN-conv & 0.598/0.599 & 2 & $72.70\pm2.83$ & $75.22\pm0.72$ & $63.92\pm1.93$ & $75.12\pm0.73$ \\
C100 conv wide & 0.584/0.597 & 2 & $47.68\pm0.62$ & $47.52\pm0.62$ & $46.01\pm2.01$ & $47.66\pm0.85$ \\
C100 conv narrow & 0.587/0.597 & 2 & $40.85\pm0.50$ & $41.47\pm0.54$ & $41.07\pm0.97$ & $41.55\pm0.39$ \\
Dry Bean & 0.373/0.400 & 3 & $86.04\pm3.02$ & $80.77\pm5.40$ & $63.04\pm9.48$ & $90.37\pm1.25$ \\
JSC & 0.329/0.433 & 25 & $76.31\pm0.02$ & $76.24\pm0.03$ & $75.51\pm0.54$ & $76.42\pm0.07$ \\
\bottomrule
\end{tabular}%
}
\end{table*}

\subsection{Structured Frontiers}
\label{app:structured_results}

\begin{table*}[!ht]
\centering
\caption{\textbf{Representative in-training structured operating points} (accuracy in \%, three seeds).}
\label{tab:app_structured}
\resizebox{0.60\textwidth}{!}{%
\begin{tabular}{llrr}
\toprule
Dataset/model & Structured setting & Cost & Accuracy \\
\midrule
MNIST EffKAN & shared-$k=6$, input & 0.7750 & $97.26\pm0.29$ \\
MNIST EffKAN & shared-$k=4$, input & 0.5534 & $97.14\pm0.11$ \\
MNIST EffKAN & neuron keep 0.75 & 0.7238 & $96.17\pm0.30$ \\
MNIST EffKAN & neuron keep 0.50 & 0.4651 & $95.69\pm0.20$ \\
MNIST KAGN & shared-$k=3$, input & 0.8000 & $97.73\pm0.05$ \\
MNIST KAGN & shared-$k=2$, input & 0.6000 & $98.20\pm0.21$ \\
MNIST KAGN & neuron keep 0.50 & 0.4651 & $97.70\pm0.15$ \\
C10 KAGN-conv & shared-$k=3$, input & 0.7993 & $80.67\pm0.42$ \\
C10 KAGN-conv & shared-$k=2$, input & 0.5995 & $75.32\pm4.47$ \\
C10 KAGN-conv & neuron keep 0.75 & 0.5826 & $78.65\pm0.59$ \\
C10 KAGN-conv & neuron keep 0.50 & 0.2783 & $68.97\pm2.19$ \\
C100 KAGN-conv (wide) & shared-$k=3$, input & 0.7960 & $53.39\pm0.91$ \\
C100 KAGN-conv (wide) & shared-$k=2$, input & 0.5972 & $52.04\pm0.75$ \\
\bottomrule
\end{tabular}%
}
\end{table*}

The three-seed training-time-versus-post-hoc comparison initially suggested advantages of $+0.42$, $+1.94$, and $+4.77$ points on MNIST KAGN, CIFAR-10 KAGN-conv, and wide CIFAR-100 KAGN-conv. The subsequent five-seed study places these timing differences within the observed seed variation. Accordingly, training-time hardening is treated as an adaptation option rather than an accuracy contribution.

\subsection{Structural Composition}
\label{app:composition}

\begin{table}[!ht]
\centering
\caption{\textbf{Composition of the two structural cost factors.}}
\label{tab:app_composition}
\resizebox{\columnwidth}{!}{%
\begin{tabular}{lrrrr}
\toprule
Model & Unit cost & Term factor & Pred. & Meas. \\
\midrule
MNIST EffKAN & 0.4651 & 0.556 & 0.2585 & 0.2584 \\
C10 KAGN-conv & 0.2783 & 0.602 & 0.1675 & 0.1676 \\
\bottomrule
\end{tabular}%
}
\end{table}

The corresponding composed accuracies are $96.18\pm0.43\%$ for MNIST and $67.96\pm1.19\%$ for CIFAR-10. The key observation is the cost factorization; accuracy itself is not assumed to factorize.

\subsection{Complete Physical-Compaction Results}
\label{app:compaction_results}

\begin{table*}[!ht]
\centering
\caption{\textbf{Physical compact checkpoints.} Sizes assume FP32 storage. All parameter reductions are measured from the actual compact checkpoint.}
\label{tab:app_compact}
\resizebox{0.9\textwidth}{!}{%
\begin{tabular}{llrrrrr}
\toprule
Dataset & Compact point & Dense params & Compact params & Reduction & Size MB & Accuracy \\
\midrule
MNIST & EffKAN sk4+n0.5 & 4{,}460{,}288 & 1{,}204{,}964 & $72.98\%$ & 4.820 & $97.98\pm0.12$ \\
MNIST & EffKAN sk4+n0.7 & 4{,}460{,}288 & 1{,}735{,}054 & $61.10\%$ & 6.940 & $97.94\pm0.13$ \\
MNIST & EffKAN sk6+n0.5 & 4{,}460{,}288 & 1{,}643{,}652 & $63.15\%$ & 6.575 & $97.87\pm0.20$ \\
MNIST & KAGN sk2+n0.5 & 2{,}583{,}072 & 766{,}656 & $70.32\%$ & 3.067 & $97.97\pm0.63$ \\
MNIST & KAGN sk2+n0.7 & 2{,}583{,}072 & 1{,}104{,}098 & $57.26\%$ & 4.416 & $98.37\pm0.07$ \\
C10 & Conv sk3+n0.7 & 592{,}932 & 238{,}386 & $59.80\%$ & 0.954 & $78.81\pm0.33$ \\
C10 & Conv sk3+n0.5 & 592{,}932 & 123{,}501 & $79.17\%$ & 0.494 & $67.48\pm1.03$ \\
C10 & Conv sk2+n0.7 & 592{,}932 & 180{,}260 & $69.60\%$ & 0.721 & $72.67\pm4.32$ \\
C10 & Conv sk2+n0.5 & 592{,}932 & 93{,}418 & $84.24\%$ & 0.374 & $64.45\pm3.13$ \\
C100 & Conv sk3+n0.7 & 2{,}586{,}328 & 1{,}075{,}542 & $58.41\%$ & 4.302 & $51.45\pm1.19$ \\
C100 & Conv sk3+n0.5 & 2{,}586{,}328 & 581{,}217 & $77.53\%$ & 2.325 & $45.41\pm1.33$ \\
C100 & Conv sk2+n0.7 & 2{,}586{,}328 & 816{,}180 & $68.44\%$ & 3.265 & $50.54\pm1.13$ \\
C100 & Conv sk2+n0.5 & 2{,}586{,}328 & 441{,}694 & $82.92\%$ & 1.767 & $45.45\pm1.45$ \\
\bottomrule
\end{tabular}%
}
\end{table*}

For CIFAR-10, extending recovery to 40 epochs raises sk3+n0.5 from $67.48\%$ to $71.13\%$ and sk2+n0.5 from $64.45\%$ to $69.73\%$ without changing model size.

Across 54 compacted checkpoints, prediction agreement with their masked source models is at least $0.9997$. Small residual logit differences remain because weakly active structure is physically removed, so compaction is described as prediction-preserving rather than bit-exact.

\subsection{Fixed-Bit QAT Frontier}
\label{app:quant_results}

\begin{table*}[!ht]
\centering
\caption{\textbf{Fixed-bit QAT grid} (accuracy in \%, three seeds). The final column is joint bit-cost.}
\label{tab:app_quant}
\resizebox{0.8\textwidth}{!}{%
\begin{tabular}{llrrrr}
\toprule
Dataset & Structure & FP32 & 8-bit & 4-bit & 4-bit cost \\
\midrule
MNIST & dense & $97.07\pm0.24$ & $97.01\pm0.23$ & $96.90\pm0.41$ & 0.1250 \\
MNIST & shared-$k=6$ & $97.26\pm0.29$ & $97.24\pm0.12$ & $97.14\pm0.10$ & 0.0969 \\
MNIST & neuron 0.5 & $95.69\pm0.20$ & $96.00\pm0.17$ & $95.26\pm0.07$ & 0.0581 \\
MNIST & top-$k=4$ & $97.58\pm0.15$ & $97.58\pm0.21$ & $97.25\pm0.13$ & 0.0692 \\
MNIST & tk4+n0.5 & $96.18\pm0.43$ & $96.42\pm0.15$ & $96.31\pm0.16$ & 0.0323 \\
C10 & dense & $81.41\pm0.58$ & $80.89\pm0.73$ & $79.56\pm0.55$ & 0.1249 \\
C10 & shared-$k=3$ & $80.67\pm0.42$ & $80.88\pm0.08$ & $78.94\pm0.88$ & 0.0999 \\
C10 & neuron 0.5 & $68.97\pm2.19$ & $68.59\pm1.53$ & $67.58\pm2.06$ & 0.0348 \\
C10 & top-$k=2$ & $77.16\pm4.09$ & $77.62\pm4.57$ & $73.46\pm4.17$ & 0.0749 \\
C10 & tk2+n0.5 & $67.96\pm1.19$ & $67.93\pm1.10$ & $66.48\pm2.21$ & 0.0209 \\
C100 narrow & dense & $49.24\pm0.69$ & $48.98\pm0.54$ & $46.86\pm0.76$ & 0.1249 \\
C100 narrow & shared-$k=3$ & $49.11\pm0.94$ & $49.06\pm0.97$ & $46.23\pm1.72$ & 0.0994 \\
C100 narrow & neuron 0.5 & $38.72\pm0.44$ & $39.41\pm1.22$ & $35.83\pm1.50$ & 0.0473 \\
C100 narrow & top-$k=2$ & -- & $48.52\pm1.18$ & $45.76\pm0.71$ & 0.0746 \\
\bottomrule
\end{tabular}%
}
\end{table*}

\subsection{Joint QAS versus Two-Stage Training}
\label{app:qas}

\begin{table}[!ht]
\centering
\caption{\textbf{Joint QAS versus budget-matched two-stage QAT at 4 bits} (five paired seeds, accuracy in \%). The paired confidence intervals include zero, so no mean-accuracy superiority is claimed.}
\label{tab:app_qas}
\resizebox{\columnwidth}{!}{%
\begin{tabular}{lrrr}
\toprule
Dataset & Joint & Two-stage & Paired $\Delta$ \\
\midrule
MNIST & $97.08\pm0.14$ & $97.14\pm0.08$ & $-0.06\pm0.14$ \\
CIFAR-10 & $79.72\pm0.30$ & $75.26\pm4.19$ & $+4.45\pm4.17$ \\
CIFAR-100 & $51.40\pm0.96$ & $50.23\pm0.84$ & $+1.17\pm1.36$ \\
\bottomrule
\end{tabular}%
}
\end{table}

On CIFAR-10, joint training has standard deviation $0.30$ compared with $4.19$ for the matched two-stage schedule. However, extending the recovery budget of the two-stage model reduces the joint-minus-two-stage differences to approximately $-0.12$, $+2.66$, and $-0.27$ points on MNIST, CIFAR-10, and CIFAR-100. Thus the earlier three-seed $+5.69$-point observation is not treated as a headline contribution.

\subsection{Dense PTQ Reference}
\label{app:ptq}

\begin{table}[!ht]
\centering
\caption{\textbf{Dense post-training quantization control} (accuracy in \%, five seeds).}
\label{tab:app_ptq}
\resizebox{0.8\columnwidth}{!}{%
\begin{tabular}{lrr}
\toprule
Dataset & PTQ 8-bit & PTQ 4-bit \\
\midrule
MNIST & $97.52\pm0.19$ & $97.15\pm0.18$ \\
CIFAR-10 & $81.87\pm0.37$ & $50.19\pm11.78$ \\
CIFAR-100 & $54.09\pm0.34$ & $9.04\pm2.43$ \\
\bottomrule
\end{tabular}%
}
\end{table}

\subsection{Learned Layer-Wise Precision}
\label{app:learned_bits}

\begin{table}[!ht]
\centering
\caption{\textbf{Learned layer-wise precision} on MNIST EfficientKAN, single seed.}
\label{tab:app_learned_bits}
\resizebox{0.7\columnwidth}{!}{%
\begin{tabular}{rrrr}
\toprule
$\lambda_{\mathrm{bits}}$ & Hardened bits & Bit-cost & Acc. \\
\midrule
$10^{-2}$ & $(6,7,8)$ & 0.9954 & 97.08 \\
$10^{-1}$ & $(3,5,7)$ & 0.4820 & 97.36 \\
$1$ & $(2,3,6)$ & 0.1007 & 97.13 \\
\bottomrule
\end{tabular}%
}
\end{table}

The experiment demonstrates that the formulation can learn heterogeneous precision, but the single-seed setting is insufficient for a primary claim.

\subsection{Cost-Model Validation}
\label{app:cost_validation}

The analytical active-cost model is intended to rank structures rather than predict exact cycle counts. Fitting effective component weights from measured latency produces base-to-term relative costs spanning approximately $0.86$--$2.16$ across families. Re-costing 683 completed operating points with these measured weights preserves their ranking: Spearman correlation lies between $0.936$ and $1.000$ for every non-degenerate family with more than three points. CIFAR-100 KAGN-conv gives approximately $0.99997$ across 177 runs, while CIFAR-10 KAGN-conv gives approximately $0.9998$ across 181 runs.

The regularization sweep also confirms that the near-dense Stage~1 point is a schedule choice. Across the tested grid, cost falls from approximately $0.996$ to $0.505$ while accuracy stays within roughly $0.3$ points.

\subsection{CUDA Latency}
\label{app:latency}

Masked models retain the dense tensor shape and show no meaningful speedup: their measured ratios remain approximately $0.95$--$1.03$ across batch sizes. Table~\ref{tab:app_latency} gives the complete batch-size sweep underlying the representative CUDA results reported in Table~\ref{tab:cuda_main} of the main paper.

\begin{table}[!ht]
\centering
\caption{\textbf{Compact/dense CUDA latency ratio.} Values below one indicate an execution speedup.}
\label{tab:app_latency}
\resizebox{\columnwidth}{!}{%
\begin{tabular}{lrrrr}
\toprule
Compact point & $b=1$ & $b=32$ & $b=256$ & $b=1024$ \\
\midrule
C10 sk3+n0.7 & 1.14 & 1.13 & 0.79 & 0.76 \\
C10 sk3+n0.5 & 1.05 & 1.04 & 0.54 & 0.54 \\
C10 sk2+n0.5 & 1.07 & 1.04 & 0.51 & 0.52 \\
C100 sk3+n0.7 & 1.15 & 1.13 & 0.75 & 0.76 \\
C100 sk2+n0.5 & 1.06 & 1.02 & 0.53 & 0.51 \\
MNIST EffKAN sk4+n0.5 & 1.04 & 1.03 & 1.03 & 0.82 \\
MNIST KAGN sk2+n0.5 & 1.09 & 1.07 & 1.07 & 0.96 \\
\bottomrule
\end{tabular}%
}
\end{table}

The difference between parameter reduction and latency reduction exposes the basis-construction residual. For example, CIFAR-10 sk3+n0.5 retains only approximately $20.8\%$ of parameters but still requires approximately $54\%$ of dense latency at batch 1024.

\subsection{Supporting Tabular Results}
\label{app:tabular}

\begin{table*}[!ht]
\centering
\caption{\textbf{Supporting tabular/time-series experiments.} Traffic uses a standardized target and is not directly comparable to published numbers on a differently scaled target. Wine has a 36-example test split and is not used for a primary claim.}
\label{tab:app_tabular}
\resizebox{0.8\textwidth}{!}{%
\begin{tabular}{lrrrrr}
\toprule
Dataset & Dense & Sparse & Cost & 8-bit & 4-bit \\
\midrule
Wine (acc. \%) & $93.52\pm1.60$ & $93.52\pm1.60$ & 1.000 & $93.52\pm1.60$ & $93.52\pm1.60$ \\
Dry Bean (acc. \%) & $92.07\pm0.13$ & $92.08\pm0.18$ & 0.647 & $92.13\pm0.15$ & $89.47\pm2.09$ \\
JSC (acc. \%) & $76.57\pm0.07$ & $76.59\pm0.08$ & 0.640 & $76.52\pm0.03$ & $74.63\pm0.22$ \\
Traffic (RMSE) & $1.0233\pm0.0043$ & $1.0252\pm0.0121$ & 0.992 & $1.0534\pm0.0201$ & $1.8693\pm0.0151$ \\
\bottomrule
\end{tabular}%
}
\end{table*}

\subsection{Additional Supporting Ablations}
\label{app:supporting}

\paragraph{Spatial kernels.}
On MNIST KAGN-conv, replacing the full spatial basis kernel with box mode reduces accuracy from $97.59\pm0.25\%$ to $55.40\%$ while reducing the parameter count from 33{,}120 to 9{,}456. EfficientKAN-conv box mode reaches $65.16\%$. These single-seed controls demonstrate that convolutional KAN compression must preserve genuine spatial structure.

\paragraph{Scale.}
A 38.37M-parameter Eight-KAGN-v2 model trains and sparsifies successfully on CIFAR-100, reaching $50.83\pm0.46\%$. It underperforms the 2.59M-parameter wide KAGN-conv anchor, so the experiment is used only as evidence that the training pipeline executes at the larger scale.

\paragraph{Symbolic regression.}
On Feynman I.6.2, spline PyKAN RMSE changes from approximately $0.0001$ dense to $0.0609\pm0.0165$ after sparsification and $0.0885\pm0.0097$ at 4 bits. The corresponding KAGN experiment is non-informative because its polynomial branch is removed entirely and the base branch already matches the dense solution.

\section{Complete FPGA/HLS Evaluation}
\label{app:fpga_full}

\subsection{Platform and Implementation}
\label{app:fpga_setup}

All hardware designs target the Xilinx ZCU104 (\texttt{xczu7ev-ffvc1156-2-e}) with Vitis HLS 2025.2.1. The clock target is 5.00\,ns with 12\% clock uncertainty. The available resources are 230{,}400 LUTs, 460{,}800 FFs, 1{,}728 DSPs, and 624 BRAM-18K blocks.

Within each KAN family, part, target clock, interface, storage strategy, and HLS pragma policy are fixed across all six designs. Only model dimensions and arithmetic precision change.

After structural hardening, gate values are folded into the exported tensors. The accelerator therefore contains no gate evaluation, mask multiplication, or runtime zero-skipping logic. Neuron/channel pruning changes compile-time input/output dimensions, while shared-$k$ compression changes the basis-loop bound and stores the original basis identities in \texttt{kept\_term\_idx}.

The EfficientKAN and GRAM MLPs are too large to store their dense weights in BRAM and therefore stream the packed weight tensors through AXI-MM from DDR. All rungs within each MLP family use the same storage architecture. KAGN-conv has a substantially smaller weight footprint and stores weights in on-chip ROM.

For the quantized MLPs, surviving low-bit weights are combined with fixed-point activation arithmetic, while the basis and normalization operations retain the precision needed by their respective implementations. The convolutional Gram basis contains transcendental operations and is not bit-identical to a purely integer golden model; this distinction affects verification tolerance but not the reported argmax accuracy.

\subsection{Complete Hardware Ladder}
\label{app:fpga_ladder}

\begin{table*}[!ht]
\centering
\caption{\textbf{Complete 18-design FPGA/HLS ladder.} Accuracy is in \% on the full MNIST test set using the exported hardware checkpoint. MLP latency is RTL co-simulation; KAGN-conv latency is the post-synthesis worst-case estimate. Resources are post-synthesis estimates.}
\label{tab:app_fpga_all}
\resizebox{\textwidth}{!}{%
\begin{tabular}{llrrrrrrrrr}
\toprule
Family & Design & Bits & Acc. & Cyc/img & Lat. ms & Img/s & LUT & FF & DSP & BRAM \\
\midrule
EffKAN & FP32 & 32 & 97.48 & 636{,}122 & 2.872 & 348 & 54{,}371 & 60{,}060 & 136 & 116 \\
EffKAN & sparse & 32 & 98.00 & 193{,}146 & 0.872 & 1{,}147 & 46{,}732 & 52{,}846 & 136 & 100 \\
EffKAN & sparse+int8 & 8 & 97.96 & 44{,}423 & 0.195 & 5{,}116 & 71{,}626 & 54{,}666 & 266 & 105 \\
EffKAN & sparse+int4 & 4 & 97.92 & 27{,}629 & 0.122 & 8{,}226 & 79{,}565 & 57{,}516 & 362 & 121 \\
EffKAN & dense int8 & 8 & 97.42 & 148{,}689 & 0.654 & 1{,}529 & 79{,}541 & 68{,}404 & 294 & 137 \\
EffKAN & dense int4 & 4 & 96.83 & 83{,}187 & 0.366 & 2{,}732 & 87{,}929 & 71{,}580 & 390 & 153 \\
\midrule
GRAM & FP32 & 32 & 98.14 & 403{,}049 & 1.898 & 527 & 36{,}189 & 39{,}630 & 70 & 109 \\
GRAM & sparse & 32 & 98.22 & 139{,}837 & 0.659 & 1{,}518 & 36{,}722 & 40{,}610 & 70 & 102 \\
GRAM & sparse+int8 & 8 & 98.11 & 32{,}109 & 0.151 & 6{,}612 & 52{,}684 & 40{,}201 & 178 & 114 \\
GRAM & sparse+int4 & 4 & 98.22 & 22{,}138 & 0.104 & 9{,}590 & 62{,}037 & 44{,}473 & 274 & 98 \\
GRAM & dense int8 & 8 & 98.14 & 91{,}731 & 0.432 & 2{,}315 & 59{,}150 & 43{,}083 & 192 & 125 \\
GRAM & dense int4 & 4 & 98.06 & 55{,}573 & 0.262 & 3{,}820 & 67{,}504 & 46{,}607 & 288 & 121 \\
\midrule
KAGN-conv & FP32 & 32 & 98.45 & 19{,}197{,}455 & 86.677 & 12 & 163{,}496 & 233{,}126 & 1{,}663 & 473 \\
KAGN-conv & sparse & 32 & 98.45 & 15{,}358{,}543 & 69.344 & 14 & 138{,}654 & 193{,}400 & 1{,}343 & 425 \\
KAGN-conv & sparse+int8 & 8 & 98.45 & 12{,}673{,}104 & 57.219 & 17 & 80{,}562 & 78{,}464 & 602 & 294 \\
KAGN-conv & sparse+int4 & 4 & 98.36 & 12{,}673{,}104 & 57.219 & 17 & 80{,}146 & 76{,}332 & 502 & 274 \\
KAGN-conv & dense int8 & 8 & 98.58 & 16{,}511{,}599 & 74.550 & 13 & 84{,}085 & 81{,}646 & 666 & 307 \\
KAGN-conv & dense int4 & 4 & 98.27 & 16{,}511{,}599 & 74.550 & 13 & 83{,}645 & 79{,}004 & 490 & 283 \\
\bottomrule
\end{tabular}%
}
\end{table*}

\subsection{Hardware Gain Decomposition}
\label{app:fpga_gains}

For EfficientKAN, structured compaction alone gives a $3.29\times$ latency reduction. Sparse+int8 reaches $14.69\times$, and sparse+int4 reaches $23.63\times$. Dense int4 reaches only $7.85\times$. At int8, the stagewise factors satisfy $3.29\times4.46\approx14.69$. The weight stream correspondingly decreases from approximately 8.45\,MB/image for dense FP32 to 0.27\,MB/image for sparse int4.

For GRAM, the analogous reductions are $2.88\times$ for sparse FP32, $12.55\times$ for sparse int8, and $18.21\times$ for sparse int4. Dense int4 gives $7.25\times$, while weight traffic falls from approximately 4.70 to 0.16\,MB/image.

The integer MLP implementations use more arithmetic resources than the FP32 baseline in some configurations because the HLS design spends additional parallel resources to minimize latency. Hence resource reduction is not the appropriate interpretation for these MLPs; their principal benefits are latency and memory traffic.

For KAGN-conv, sparse+int4 reaches only $1.51\times$ serial latency reduction but reduces DSP count by $3.31\times$, FF count by approximately $3.05\times$, LUT count by approximately $2.04\times$, and BRAM use by $1.73\times$. It is therefore primarily a capacity result.

\subsection{Device Utilization}
\label{app:fpga_resources}

\begin{table*}[!ht]
\centering
\caption{\textbf{Post-synthesis resource utilization} as a percentage of the ZU7EV device budget.}
\label{tab:app_fpga_resource}
\resizebox{0.6\textwidth}{!}{%
\begin{tabular}{llrrrr}
\toprule
Family & Design & LUT & FF & DSP & BRAM \\
\midrule
EffKAN & FP32 & $23.6\%$ & $13.0\%$ & $7.9\%$ & $18.6\%$ \\
EffKAN & sparse & $20.3\%$ & $11.5\%$ & $7.9\%$ & $16.0\%$ \\
EffKAN & sparse+int8 & $31.1\%$ & $11.9\%$ & $15.4\%$ & $16.8\%$ \\
EffKAN & sparse+int4 & $34.5\%$ & $12.5\%$ & $20.9\%$ & $19.4\%$ \\
EffKAN & dense int8 & $34.5\%$ & $14.8\%$ & $17.0\%$ & $22.0\%$ \\
EffKAN & dense int4 & $38.2\%$ & $15.5\%$ & $22.6\%$ & $24.5\%$ \\
\midrule
GRAM & FP32 & $15.7\%$ & $8.6\%$ & $4.1\%$ & $17.5\%$ \\
GRAM & sparse & $15.9\%$ & $8.8\%$ & $4.1\%$ & $16.3\%$ \\
GRAM & sparse+int8 & $22.9\%$ & $8.7\%$ & $10.3\%$ & $18.3\%$ \\
GRAM & sparse+int4 & $26.9\%$ & $9.7\%$ & $15.9\%$ & $15.7\%$ \\
GRAM & dense int8 & $25.7\%$ & $9.3\%$ & $11.1\%$ & $20.0\%$ \\
GRAM & dense int4 & $29.3\%$ & $10.1\%$ & $16.7\%$ & $19.4\%$ \\
\midrule
KAGN-conv & FP32 & $71.0\%$ & $50.6\%$ & $96.2\%$ & $75.8\%$ \\
KAGN-conv & sparse & $60.2\%$ & $42.0\%$ & $77.7\%$ & $68.1\%$ \\
KAGN-conv & sparse+int8 & $35.0\%$ & $17.0\%$ & $34.8\%$ & $47.1\%$ \\
KAGN-conv & sparse+int4 & $34.8\%$ & $16.6\%$ & $29.1\%$ & $43.9\%$ \\
KAGN-conv & dense int8 & $36.5\%$ & $17.7\%$ & $38.5\%$ & $49.2\%$ \\
KAGN-conv & dense int4 & $36.3\%$ & $17.1\%$ & $28.4\%$ & $45.4\%$ \\
\bottomrule
\end{tabular}%
}
\end{table*}

\subsection{Functional Verification}
\label{app:fpga_verify}

Every exported hardware design is checked against a NumPy golden reference. All twelve MLP configurations pass RTL co-simulation. Their checked predictions agree with the golden reference by argmax. Integer EfficientKAN implementations using the identical integer arithmetic are bit-exact on the 1000-image verification set. Floating-point variants exhibit only small logit differences from reassociation, with observed maximum errors on the order of $10^{-5}$.

Full-image RTL co-simulation is infeasible for KAGN-conv because one inference requires approximately 12--19 million cycles. Each of the six convolutional designs is therefore checked using a 64-image C-simulation spot test, and all six obtain 64/64 argmax agreement.

The quantized KAGN-conv kernels evaluate the Gram basis using floating-point transcendental operations, so their logits are not expected to be bit-exact to a purely integer golden reference. One testbench therefore emits a strict bit-level failure despite perfect argmax agreement; this is a tolerance artifact rather than a classification failure.

\subsection{Scope of the Hardware Evidence}
\label{app:fpga_scope}

The FPGA study intentionally separates two deployment regimes. For the DDR-streaming MLPs, structural compaction and low precision both shrink the dominant weight-transfer workload and therefore translate into large batch-1 latency reductions.

For KAGN-conv, the one-MAC-per-cycle implementation remains dominated by spatial convolution and basis evaluation. Compression instead frees device resources. The resulting headroom could be used to replicate multiple compressed engines or to deploy a larger model, but such replication is not implemented or measured here.

Accordingly, we describe the results as FPGA/HLS realization evidence. They are not board-level measurements of wall-clock latency, energy, or power. MLP latency comes from RTL co-simulation and convolutional latency from HLS synthesis estimates, as labeled throughout.

\end{document}